\documentclass[10pt,journal,compsoc]{IEEEtran}

\ifCLASSOPTIONcompsoc
  \usepackage{cite}
\else
\fi
\ifCLASSINFOpdf
\else
\fi
\newcommand\MYhyperrefoptions{bookmarks=true,bookmarksnumbered=true,
pdfpagemode={UseOutlines},plainpages=false,pdfpagelabels=true,
colorlinks=true,linkcolor={black},citecolor={black},urlcolor={black},
pdftitle={A Review on Deep Learning Techniques for Video Prediction},
pdfsubject={Typesetting},
pdfauthor={Sergiu Oprea},
pdfkeywords={Video prediction review, future frame prediction, deep learning, representation learning, self-supervised learining}}
\ifCLASSINFOpdf
\usepackage[\MYhyperrefoptions,pdftex]{hyperref}
\else
\usepackage[\MYhyperrefoptions,breaklinks=true,dvips]{hyperref}
\usepackage{breakurl}
\fi
\usepackage{cite}
\usepackage{graphicx}
\usepackage{url}
\usepackage{xurl}
\usepackage{xcolor}
\usepackage{amssymb}
\usepackage[acronym]{glossaries}
\setkeys{glslink}{hyper=false}

\usepackage{siunitx}

\newacronym{cnn}{CNN}{Convolutional Neural Network}
\newacronym{decnn}{deCNN}{Deconvolutional Neural Network}
\newacronym{lstm}{LSTM}{Long Short-Term Memory}
\newacronym{rnn}{RNN}{Recurrent Neural Network}
\newacronym{gdl}{GDL}{Gradient Difference Loss}
\newacronym{mse}{MSE}{Mean-Squared Error}
\newacronym{lapgan}{LAPGAN}{Laplacian Generative Adversarial Network}
\newacronym{pgn}{PGN}{Predictive Generative Network}
\newacronym{advloss}{AL}{adversarial loss}
\newacronym{gan}{GAN}{Generative Adversarial Network}
\newacronym{cgan}{cGAN}{conditional Generative Adversarial Network}
\newacronym{gru}{GRU}{Gated Recurrent Unit}
\newacronym{convlstm}{ConvLSTM}{convolutional LSTM}
\newacronym{prednet}{PredNet}{Predictive Coding Network}
\newacronym{3d}{3d}{three-dimensional}
\newacronym{2d}{2d}{two-dimensional}
\newacronym{rl}{RL}{reinforcement learning}
\newacronym{pgp}{PGP}{Predictive Gating Pyramid}
\newacronym{gae}{GAE}{Gated Autoencoder}
\newacronym{vstm}{VSTM}{Visual Short-term Memory}
\newacronym{dct}{DCT}{Discrete Cosine Transform}
\newacronym{st}{ST}{Spatial Transformer}
\newacronym{stp}{STP}{Spatial Transformer Predictor}
\newacronym{dna}{DNA}{Dynamic Neural Advection}
\newacronym{cdna}{CDNA}{Convolutional Dynamic Neural Advection}
\newacronym{dfn}{DFN}{Dynamic Filter Networks}
\newacronym{dvf}{DVF}{Deep Voxel Flow}
\newacronym{tv}{TV}{Total Variation}
\newacronym{vae}{VAE}{Variational Autoencoder}
\newacronym{mcnet}{MCnet}{Motion-content Network}
\newacronym{lad}{LAD}{least absolute deviation}
\newacronym{ls}{LS}{least square error}
\newacronym{ae}{AE}{Autoencoder}
\newacronym{ed}{ED}{Encoder-decoder}
\newacronym{pearl}{PEARL}{Parsing with prEdictive feAtuRe Learning}
\newacronym{dcl}{DCL}{Dynamic Convolutional Layer}
\newacronym{drnet}{DRNET}{Disentangled-representation Net}
\newacronym{frnn}{fRNN}{folded Recurrent Neural Network}
\newacronym{fstn}{FSTN}{Flexible Spatio-semporal Network}
\newacronym{cvae}{cVAE}{conditioned Variational Autoencoder}
\newacronym{svg}{SVG}{Stochastic Video Generation}
\newacronym{ce}{CE}{Cross Entropy}
\newacronym{een}{EEN}{Error Encoding Network}
\newacronym{savp}{SAVP}{Stochastic Adversarial Video Prediction}
\newacronym{ile}{ILE}{Invertible Linear Embedding}
\newacronym{vrnn}{VRNN}{Variational Recurrent Neural Network}
\newacronym{epva}{EPVA}{EPVA}
\newacronym{log}{LoG}{Laplacian of Gaussians}
\newacronym{mdlstm}{MD-LSTM}{multidimensional LSTM}
\newacronym{ddpae}{DDPAE}{Decompositional Disentangled Predictive Autoencoder}
\newacronym{sv2p}{SV2P}{Stochastic Variational Video Prediction}
\newacronym{e3dlstm}{E3d-LSTM}{\textit{eidetic} 3d LSTM}
\newacronym{pggan}{PGGAN}{Progressively Growing GAN}
\newacronym{wgangp}{WGAN-GP}{Wasserstein GAN with gradient penalty}
\newacronym{svm}{SVM}{Support Vector Machine}
\newacronym{fps}{fps}{frames per second}
\newacronym{fvbn}{FVBN}{Fully Visible Belief Network}
\newacronym{bptt}{BPTT}{Backpropagation through time}
\newacronym{lvae}{LVAE}{Ladder Variational Autoencoder}
\newacronym{ssim}{SSIM}{Structural Similarity Index Measure}
\newacronym{psnr}{PSNR}{Peak Signal to Noise Ratio}
\newacronym{fvd}{FVD}{Fréchet Video Distance}
\newacronym{lpips}{LPIPS}{Learned Perceptual Image Patch Similarity}
\newacronym{iou}{IoU}{Intersection over Union}
\newacronym{rln}{RLN}{Recurrent Ladder Network}
\newacronym{vpn}{VPN}{Video Pixel Network}
\newacronym{vln}{VLN}{Video Ladder Network}
\newacronym{amt}{AMT}{Amazon Mechanical Turk}
\newacronym{ved}{VED}{Variational Encoder-Decoder}
\newacronym{tsru}{TSRU}{Transformation-based Spatial Recurrent Unit}
\newacronym{trivdganfp}{TrIVD-GAN-FP}{Transformation-based \& TrIple Video Discriminator GAN}
\newacronym{hvs}{HVS}{Human Visual System}
\newacronym{sdc}{SDC}{Spatially Displaced Convolution}
\newacronym{crevnet}{CrevNet}{Conditionally Reversible Network}
\newacronym{rpm}{RPM}{Reversible Predictive Model}

\usepackage{tikz}
\usetikzlibrary{shapes}
\usetikzlibrary{arrows}
\usetikzlibrary{positioning}
\usepackage{array}
\usepackage{setspace}
\usepackage{lipsum}

\newcolumntype{M}[1]{>{\centering\arraybackslash}m{#1}}

\usepackage{booktabs}
\usepackage{pifont}
\newcommand{\ra}[1]{\renewcommand{\arraystretch}{#1}}
\newcommand{\0}{\mathord{\mspace{2mu}\star\mspace{2mu}}}
\renewcommand{\tabcolsep}{5pt}
\newcommand{\xmark}{\ding{55}}
\usepackage[para,online,flushleft]{threeparttable}

\usepackage{subcaption}
\usepackage{forest}
\usetikzlibrary{arrows.meta,angles,quotes}
\newcommand{\comment}[1]{}

\begin{document}
%
\title{A Review on Deep Learning Techniques for Video Prediction}
%
%
%
%

\author{S.~Oprea,
		P.~Martinez-Gonzalez,
		A.~Garcia-Garcia,
		J.A.~Castro-Vargas,
        S.~Orts-Escolano,
        J.~Garcia-Rodriguez,
        and~A.~Argyros
\IEEEcompsocitemizethanks{\IEEEcompsocthanksitem S.~Oprea, P.~Martinez-Gonzalez A.~Garcia-Garcia, J.~A.~Castro-Vargas, and J.~Garcia-Rodriguez are with the 3D Perception Lab (3DPL), Department of Computer Technology, University of Alicante, Carrer de San Vicente del Raspeig s/n, E-03690 San Vicent del Raspeig Spain, Spain. \protect\\
E-mail: \{soprea, pmartinez, jacastro, jgarcia\}@dtic.ua.es
\IEEEcompsocthanksitem A.~Garcia-Garcia is with the Institute of Space Sciences (ICE-CSIC), Campus UAB, Carrer de Can Magrans s/n, E-08193 Barcelona, Spain. \protect\\
E-mail: garciagarcia@ice.csic.es.
\IEEEcompsocthanksitem S. Orts-Escolano is with the Department of Computer Science and Artificial Intelligence (DCCIA), University of Alicante, Carrer de San Vicente del Raspeig s/n, E-03690 San Vicent del Raspeig Spain, Spain. \protect\\
E-mail: sorts@dccia.ua.es.
\IEEEcompsocthanksitem A. Argyros is with the Institute of Computer Science, FORTH, Heraklion GR-700 13, Greece and with the Computer Science Department, University of Crete, Heraklion, Rethimno 741 00, Greece. \protect\\ E-mail: argyros@ics.forth.gr.}}
\IEEEtitleabstractindextext{%
\begin{abstract}
The ability to predict, anticipate and reason about future outcomes is a key component of intelligent decision-making systems. In light of the success of deep learning in computer vision, deep-learning-based video prediction emerged as a promising research direction. Defined as a self-supervised learning task, video prediction represents a suitable framework for representation learning, as it demonstrated potential capabilities for extracting meaningful representations of the underlying patterns in natural videos. Motivated by the increasing interest in this task, we provide a review on the deep learning methods for prediction in video sequences. We firstly define the video prediction fundamentals, as well as mandatory background concepts and the most used datasets. Next, we carefully analyze existing video prediction models organized according to a proposed taxonomy, highlighting their contributions and their significance in the field. The summary of the datasets and methods is accompanied with experimental results that facilitate the assessment of the state of the art on a quantitative basis. The paper is summarized by drawing some general conclusions, identifying open research challenges and by pointing out future research directions.    
\end{abstract}

}

\maketitle

\IEEEdisplaynontitleabstractindextext

%
\IEEEpeerreviewmaketitle

\section{Introduction}
\label{sec:introduction}
\begin{figure}[!tb]
	\centering
	\resizebox{\linewidth}{!}{
		\input{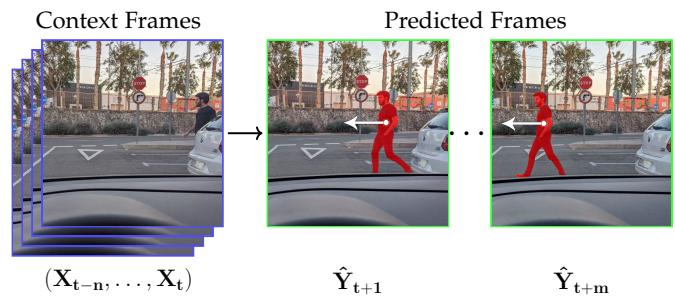}}
	\caption{A pedestrian appeared from behind the white car with the intention of crossing the street. The autonomous car must make a call: hit the emergency braking routine or not. This all comes down to predict the next frames~$(\hat{Y}_{t+1},\ldots,\hat{Y}_{t+m})$ given a sequence of context frames~$(X_{t-n}, \ldots, X_{t})$, where $n$ and $m$ denote the number of context and predicted frames, respectively. From these predictions at a representation level (RGB, high-level semantics, etc.) a decision-making system would make the car avoid the collision.}
	\label{fig:pedestrian}
\end{figure}

\IEEEPARstart{W}{ill} the car hit the pedestrian? That might be one of the questions that comes to our minds when we observe Figure \ref{fig:pedestrian}. Answering this question might be in principle a hard task; however, if we take a careful look into the image sequence we may notice subtle clues that can help us predicting into the future, e.g., the person's body indicates that he is running fast enough so he will be able to escape the car's trajectory. This example is just one situation among many others in which predicting future frames in video is useful.

In general terms, the prediction and anticipation of future events is a key component of intelligent decision-making systems. Despite the fact that we, humans, solve this problem quite easily and effortlessly,
it is extremely challenging from a machine's point of view.
Some of the factors that contribute to such complexity are occlusions, camera movement, lighting conditions, clutter, or object deformations. Nevertheless, despite such challenging conditions, many predictive methods have been applied with a certain degree of success in a broad range of application domains such as autonomous driving, robot navigation and human-machine interaction. 
Some of the tasks in which future prediction has been applied successfully are: anticipating activities and events \cite{Nguyen2012,Kitani2012,Vondrick2016a,Zeng2017}, long-term planning \cite{Shalev-Shwartz2016}, future prediction of object locations \cite{Makansi2019}, video interpolation \cite{Liu2017}, predicting instance/semantic segmentation maps~\cite{Luc2018,Bhattacharyya2019,Terwilliger2019}, prediction of pedestrian trajectories in traffic \cite{Bhattacharyya2018}, anomaly detection~\cite{Liu2018a}, precipitation nowcasting~\cite{Shi2015,Shi2017}, and autonomous driving~\cite{Hu2020}.

The great strides made by deep learning algorithms in a variety of research fields such as semantic segmentation~\cite{Garcia2018}, human action recognition and prediction~\cite{Kong2018}, object pose estimation~\cite{Sahin2020} and registration~\cite{Villena2020} to name a few, motivated authors to explore deep representation-learning models for future video frame prediction. What made the deep architectures take a leap over the traditional approaches is their ability to learn adequate
representations from high-dimensional data in an end-to-end fashion without hand-engineered features~\cite{LeCun2015}. Deep learning-based models fit perfectly into the learning by prediction paradigm, enabling the extraction of meaningful spatio-temporal correlations from video data in a self-supervised fashion.

In this review, we put our focus on deep learning techniques and how they have been extended or applied to future video prediction. We limit this review to the future video prediction given the  context of a sequence of previous frames, leaving aside methods that predict future from a static image. In this context, the terms
video prediction, future frame prediction, next video frame prediction, future frame forecasting, and future frame generation are used interchangeably. To the best of our knowledge, this is the first review in the literature that focuses on video prediction using deep learning techniques.

This review is organized as follows. First, Sections~\ref{sec:video_prediction} and \ref{sec:background} lay down the terminology and explain important background concepts that will be necessary throughout the rest of the paper. Next, Section~\ref{sec:datasets} surveys the datasets used by the video prediction methods that are carefully reviewed in Section~\ref{sec:prediction_methods}, providing a comprehensive description as well as an analysis of their strengths and weaknesses. Section~\ref{sec:performance_eval} analyzes typical metrics and evaluation protocols for the aforementioned methods and provides quantitative results for them in the reviewed datasets. Section \ref{sec:discussion} presents a brief discussion on the presented proposals and enumerates potential future research directions. Finally, Section \ref{sec:conclusion} summarizes the paper and draws conclusions about this work.
\section{Video Prediction}
\label{sec:video_prediction}
The ability to predict, anticipate and reason about future events is the essence of intelligence~\cite{Hawkins2004} and one of the main goals of decision-making systems. This idea has biological roots, and also draws inspiration from the predictive coding paradigm~\cite{Rao1999} borrowed from the cognitive neuroscience field~\cite{Mumford1992}. From a neuroscience perspective, the human brain builds complex mental representations of the physical and causal rules that govern the world. This is primarily through observation and interaction~\cite{Cleeremans1991,Cleeremans1993,Baker2014}. The common sense we have about the world arises from the conceptual acquisition and the accumulation of background knowledge from early ages, e.g. biological motion and intuitive physics to name a few. But how can the brain check and refine the learned mental representations from its raw sensory input? The brain is continuously learning through prediction, and refines the already understood world models from the mismatch between its predictions and what actually happened~\cite{Ouden2012}. This is the essence of the predictive coding paradigm that early works tried to implement as computational models~\cite{Softky1995,Rao1999,Deco2001,Hollingworth2004}. 

Video prediction task closely captures the fundamentals of the predictive coding paradigm and it is considered the intermediate step between raw video data and decision making. Its potential to extract meaningful representations of the underlying patterns in video data makes the video prediction task a promising avenue for self-supervised representation learning.

\subsection{Problem Definition}
We formally define the task of predicting future frames in videos, i.e. video prediction, as follows. Let ${\mathbf{X}_{t}\in\mathrm{R}^{w \times h \times c}}$ be the $t$-th frame in the video sequence ${\mathbf{X}=(X_{t-n},\ldots,X_{t-1},X_t)}$ with $n$ frames, where $w$, $h$, and $c$ denote width, height, and number of channels, respectively. The target is to predict the next frames ${\mathbf{Y}=(\hat{Y}_{t+1},\hat{Y}_{t+2},\ldots,\hat{Y}_{t+m})}$ from the input $\mathbf{X}$.

Under the assumption that good predictions can only be the result of accurate representations, learning by prediction is a feasible approach to verify how accurately the system has learned the underlying patterns in the input data. In other words, it represents a suitable framework for representation learning~\cite{Bengio2013,Wang2015}. The essence of predictive learning paradigm is the prediction of plausible future outcomes from a set of historical inputs. On this basis, the task of video prediction is defined as: given a sequence of video frames as context, predict the subsequent frames --generation of
continuing video given a sequence of previous frames. Different from video generation that is mostly unconditioned, video prediction is conditioned on a previously learned representation from a sequence of input frames. At a first glance, and in the context of learning paradigms, we can think about the future video frame prediction task as a supervised learning approach because the target frame acts as a label. However, as this information is already available in the input video sequence, no extra labels or human supervision is needed. Therefore, learning by prediction is a self-supervised task, filling the gap between supervised and unsupervised learning. 

\subsection{Exploiting the Time Dimension of Videos}
Unlike static images, videos provide complex transformations and motion patterns in the time dimension. At a fine granularity, if we focus on a small patch at the same spatial location across consecutive time steps, we could identify a wide range of local visually similar deformations due to the temporal coherence. In contrast, by looking at the big picture, consecutive frames would be visually different but semantically coherent. This variability in the visual appearance of a video at different scales is mainly due to, occlusions, changes in the lighting conditions, and camera motion, among other factors. From this source of temporally ordered visual cues, predictive models are able to extract representative spatio-temporal correlations depicting the dynamics in a video sequence. For instance, Agrawal \textit{et al.}~\cite{Agrawal2015} established a direct link between vision and motion, attempting to reduce supervision efforts when training deep predictive models.

Recent works study how important is the time dimension for video understanding models~\cite{Huang2018}. The implicit temporal ordering in videos, also known as the arrow of time, indicates whether a video sequence is playing forward or backward. This temporal direction is also used in the literature as a supervisory signal~\cite{Pickup2014,Wei2018,Misra2016}. This further encouraged predictive models to implicitly or explicitly model the spatio-temporal correlations of a video sequence to understand the dynamics of a scene. The time dimension of a video reduces the supervision effort and makes the prediction task self-supervised. 

\subsection{Dealing with Stochasticity}
\begin{figure}
	\centering
	\resizebox{\linewidth}{!}{
		\begin{tikzpicture}[]
	\tikzset{box/.style={black, draw=black, fill=none, very thick, minimum height=4 em, minimum width=16 em}}

	\foreach \k in {0,...,3}
	{
		\foreach \i in {0,...,4}
		{
			\foreach \j in {0,...,4}
			{
				\node[box, fill=gray!50, draw=gray!50, minimum height=1 em, minimum width=1 em] (l1-\i-\j-\k) at (\i em + 0.75 em * \k em, \j em) {};
			}
		}
	}

	\node[box, fill=black, draw=black, minimum height=1 em, minimum width=1 em] at (l1-2-2-0){};
	
	\node[box, fill=black, draw=black, minimum height=1 em, minimum width=1 em] at ([yshift=0.4em, xshift=0.4em]l1-2-2-1){};
	\node[box, fill=black, draw=black, minimum height=1 em, minimum width=1 em] at ([yshift=0.8em, xshift=0.8em]l1-2-2-2){};
	\node[box, fill=black, draw=black, minimum height=1 em, minimum width=1 em] at ([yshift=1.2em, xshift=1.2em]l1-2-2-3){};

	\foreach \k in {0,...,3}
	{
		\foreach \i in {0,...,4}
		{
			\foreach \j in {0,...,4}
			{
				\node[box, fill=gray!50, draw=gray!50, minimum height=1 em, minimum width=1 em] (l2-\i-\j-\k) at (\i em + 0.75 em * \k em, \j em - 6 em) {};
			}
		}
	}

	\node[box, fill=black, draw=black, minimum height=1 em, minimum width=1 em] at (l2-2-2-0){};



	\node[box, fill=black, draw=none, minimum height=1 em, minimum width=1 em, opacity=0.25] at ([yshift=-0.4em, xshift=0em]l2-2-2-1){};
	\node[box, fill=black, draw=none, minimum height=1 em, minimum width=1 em, opacity=0.25] at ([yshift=0.4em, xshift=0em]l2-2-2-1){};
	\node[box, fill=black, draw=none, minimum height=1 em, minimum width=1 em, opacity=0.25] at ([yshift=-0.4em, xshift=0.4em]l2-2-2-1){};
	\node[box, fill=black, draw=none, minimum height=1 em, minimum width=1 em, opacity=0.25] at ([yshift=0em, xshift=0.4em]l2-2-2-1){};
	\node[box, fill=black, draw=none, minimum height=1 em, minimum width=1 em, opacity=0.25] at ([yshift=0.4em, xshift=0.4em]l2-2-2-1){};
	\node[box, fill=black, draw=none, minimum height=1 em, minimum width=1 em, opacity=0.25] at ([yshift=-0.4em, xshift=-0.4em]l2-2-2-1){};
	\node[box, fill=black, draw=none, minimum height=1 em, minimum width=1 em, opacity=0.25] at ([yshift=0em, xshift=-0.4em]l2-2-2-1){};
	\node[box, fill=black, draw=none, minimum height=1 em, minimum width=1 em, opacity=0.25] at ([yshift=0.4em, xshift=-0.4em]l2-2-2-1){};

	\node[box, fill=black, draw=none, minimum height=1 em, minimum width=1 em, opacity=0.25] at ([yshift=-0.8em, xshift=0em]l2-2-2-2){};
	\node[box, fill=black, draw=none, minimum height=1 em, minimum width=1 em, opacity=0.25] at ([yshift=0.8em, xshift=0em]l2-2-2-2){};
	\node[box, fill=black, draw=none, minimum height=1 em, minimum width=1 em, opacity=0.25] at ([yshift=-0.8em, xshift=0.8em]l2-2-2-2){};
	\node[box, fill=black, draw=none, minimum height=1 em, minimum width=1 em, opacity=0.25] at ([yshift=0em, xshift=0.8em]l2-2-2-2){};
	\node[box, fill=black, draw=none, minimum height=1 em, minimum width=1 em, opacity=0.25] at ([yshift=0.8em, xshift=0.8em]l2-2-2-2){};
	\node[box, fill=black, draw=none, minimum height=1 em, minimum width=1 em, opacity=0.25] at ([yshift=-0.8em, xshift=-0.8em]l2-2-2-2){};
	\node[box, fill=black, draw=none, minimum height=1 em, minimum width=1 em, opacity=0.25] at ([yshift=0em, xshift=-0.8em]l2-2-2-2){};
	\node[box, fill=black, draw=none, minimum height=1 em, minimum width=1 em, opacity=0.25] at ([yshift=0.8em, xshift=-0.8em]l2-2-2-2){};

	\node[box, fill=black, draw=none, minimum height=1 em, minimum width=1 em, opacity=0.25] at ([yshift=-1.2em, xshift=0em]l2-2-2-3){};
	\node[box, fill=black, draw=none, minimum height=1 em, minimum width=1 em, opacity=0.25] at ([yshift=1.2em, xshift=0em]l2-2-2-3){};
	\node[box, fill=black, draw=none, minimum height=1 em, minimum width=1 em, opacity=0.25] at ([yshift=-1.2em, xshift=1.2em]l2-2-2-3){};
	\node[box, fill=black, draw=none, minimum height=1 em, minimum width=1 em, opacity=0.25] at ([yshift=0em, xshift=1.2em]l2-2-2-3){};
	\node[box, fill=black, draw=none, minimum height=1 em, minimum width=1 em, opacity=0.25] at ([yshift=1.2em, xshift=1.2em]l2-2-2-3){};
	\node[box, fill=black, draw=none, minimum height=1 em, minimum width=1 em, opacity=0.25] at ([yshift=-1.2em, xshift=-1.2em]l2-2-2-3){};
	\node[box, fill=black, draw=none, minimum height=1 em, minimum width=1 em, opacity=0.25] at ([yshift=0em, xshift=-1.2em]l2-2-2-3){};
	\node[box, fill=black, draw=none, minimum height=1 em, minimum width=1 em, opacity=0.25] at ([yshift=1.2em, xshift=-1.2em]l2-2-2-3){};

	\node at ([yshift=1.75 em, xshift=-1.75 em]l1-4-4-0) { Context Frame};
	\draw[-latex, very thick] ([yshift=1.25 em, xshift=3 em]l1-4-4-0.north) -- node [yshift=0.75 em]{Time}([yshift=1.25 em]l1-4-4-3.north);

	\node[rotate=90] at ([yshift=-2 em, xshift=-6 em]l1-4-4-0.center) {Deterministic};
	\node[rotate=90] at ([yshift=-2 em, xshift=-6 em]l2-4-4-0.center) {Probabilistic};

\end{tikzpicture}
	}
	\caption{At top, a deterministic environment where a geometric object, e.g. a black square, starts moving following a random direction. At bottom, probabilistic outcome. Darker areas correspond to higher probability outcomes. As uncertainty is introduced, probabilities get blurry and averaged. Figure inspired by \cite{Babaeizadeh2018}.}
	\label{fig:deterministic}
\end{figure}
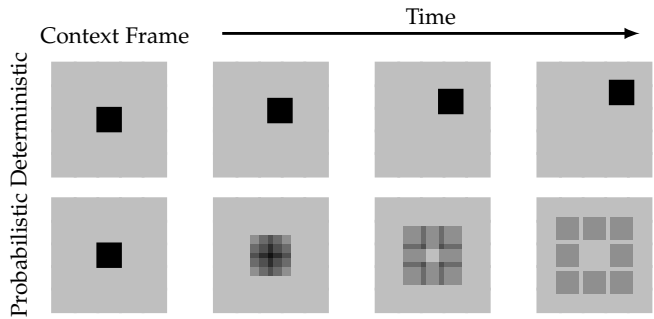

Predicting how a square is moving, could be extremely challenging even in a deterministic environment such as the one represented in Figure~\ref{fig:deterministic}. The lack of contextual information and the multiple equally probable outcomes hinder the prediction task. But, what if we use two consecutive frames as context? Under this configuration and assuming a physically perfect environment, the square will be indefinitely moving in the same direction. This represents a deterministic outcome, an assumption that many authors made in order to deal with future uncertainty. Assuming a deterministic outcome would narrow the prediction space to a unique solution. However, this assumption is not suitable for natural videos. The future is by nature multimodal, 
since the probability distribution defining all the possible future outcomes in a context has multiple modes, i.e. there are multiple equally probable and valid outcomes. Furthermore, on the basis of a deterministic universe, we indirectly assume that all possible outcomes are reflected in the input data. These assumptions make the prediction under uncertainty an extremely challenging task. 

Most of the existing deep learning-based models in the literature are deterministic. Although the future is uncertain, a deterministic prediction would suffice some easily predictable situations. For instance, most of the movement of a car is largely deterministic, while only a small part is uncertain. However, when multiple predictions are equally probable, a deterministic model will learn to average between all the possible outcomes. This unpredictability is visually represented in the predictions as blurriness, especially on long time horizons. As deterministic models are unable to handle real-world settings characterized by chaotic dynamics, authors considered that incorporating uncertainty to the model is a crucial aspect. Probabilistic approaches dealing with these issues are discussed in Section~\ref{subsec:m_uncertainty}.

\subsection{The Devil is in the Loss Function}
The design and selection of the loss function for the video prediction task is of utmost importance. Pixel-wise losses, e.g. \gls{ce}, $\ell_2$, $\ell_1$ and \gls{mse}, are widely used in both unstructured and structured predictions. Although leading to plausible predictions in deterministic scenarios, such as synthetic datasets and video games, they struggle with the inherent uncertainty of natural videos. In a probabilistic environment, with different equally probable outcomes, pixel-wise losses aim to accommodate uncertainty by blurring the prediction, as we can observe in Figure \ref{fig:deterministic}. In other words, the deterministic loss functions average out multiple equally plausible outcomes in a single, blurred prediction. In the pixel space, these losses are unstable to slight deformations and fail to capture discriminative representations to efficiently regress the broad range of possible outcomes. This makes difficult to draw predictions maintaining the consistency with our visual similarity notion. Besides video prediction, several studies analyzed the impact of different loss functions in image restoration~\cite{Zhao2017}, classification~\cite{Janocha2017}, camera pose regression~\cite{Kendall2017} and structured prediction~\cite{Hwang2019}, among others. This fosters reasoning about the importance of the loss function, particularly when making long-term predictions in high-dimensional and multimodal natural videos.

Most of distance-based loss functions, such as based on $\ell_p$ norm, come from the assumption that data is drawn from a Gaussian distribution. But, how these loss functions address multimodal distributions? Assuming that a pixel is drawn from a bimodal distribution with two equally likely modes $Mo_1$ and $Mo_2$, the mean value ${\overline{Mo}=(Mo_1 + Mo_2)/2}$ would minimize the $\ell_p$-based losses over the data, even if $\overline{Mo}$ has very low probability~\cite{Mathieu2016}. This suggests that the average of two equally probable outcomes would minimize distance-based losses such as, the \gls{mse} loss. However, this applies to a lesser extent when using $\ell_1$ norm as the pixel values would be the median of the two equally likely modes in the distribution. In contrast to the $\ell_2$ norm that emphasizes outliers with the squaring term, the $\ell_1$ promotes sparsity thus making it more suitable for prediction in high-dimensional data \cite{Mathieu2016}. Based on the $\ell_2$ norm, the \gls{mse} is also commonly used in the training of video prediction models. However, it produces low reconstruction errors by merely averaging all the possible outcomes in a blurry prediction as uncertainty is introduced. In other words, the mean image would minimize the \gls{mse} error as it is the global optimum, thus avoiding finer details such as facial features and subtle movements as they are noise for the model. Most of the video prediction approaches rely on pixel-wise loss functions, obtaining roughly accurate predictions in easily predictable datasets.

One of the ultimate goals of many video prediction approaches is to palliate the blurry predictions when it comes to uncertainty. For this purpose, authors broadly focused on: directly improving the loss functions; exploring adversarial training; alleviating the training process by reformulating the problem in a higher-level space; or exploring probabilistic alternatives. Some promising results were reported by combining the loss functions with sophisticated regularization terms, e.g. the \gls{gdl} to enhance prediction sharpness~\cite{Mathieu2016} and the \gls{tv} regularization to reduce visual artifacts and enforce coherence~\cite{Liu2017}. Perceptual losses were also used to further improve the visual quality of the predictions~\cite{Dosovitskiy2016,Johnson2016,Ledig2017,Sajjadi2017,Zhu2016}. However, in light of the success of the \glspl{gan}, adversarial training emerged as a promising alternative to disambiguate between multiple equally probable modes. It was widely used in conjunction with different distance-based losses such as: \gls{mse}~\cite{Lotter2015}, $\ell_2$~\cite{Chen2017,Jin2017,Wichers2018}, or a combination of them~\cite{Mathieu2016,Villegas2017,Walker2017,Liang2017,Luc2017,Hu2019}. To alleviate the training process, many authors reformulated the optimization process in a higher-level space (see Section \ref{subsec:m_high_level}). While great strides have been made to mitigate blurriness, most of the existing approaches still rely on distance-based loss functions. As a consequence, the regress-to-the-mean problem remains an open issue. This has further encouraged authors to reformulate existing deterministic models in a probabilistic fashion.
\section{Backbone Deep Learning Architectures}
\label{sec:background}
In this section, we will briefly review the most common deep networks that are used as building blocks for the video prediction models discussed in this review: convolutional neural networks, recurrent networks, and generative models.

\subsection{Convolutional Models}
Convolutional layers are the basic building blocks of deep learning architectures designed for visual reasoning since the \glspl{cnn} efficiently model the spatial structure of images~\cite{Lecun1998}. As we focus on the visual prediction, \glspl{cnn} represent the foundation of predictive learning literature. However, their performance is limited by the intra-frame and inter-frame dependencies.

Convolutional operations account for short-range intra-frame dependencies due to their limited receptive fields, determined by the kernel size. This is a well-addressed issue, that many authors circumvented by (1) stacking more convolutional layers~\cite{Jain2007}, (2) increasing the kernel size (although it becomes prohibitively expensive), (3) by linearly combining multiple scales~\cite{Mathieu2016} as in the reconstruction process of a Laplacian pyramid~\cite{Denton2015}, (4) using dilated convolutions to capture long-range spatial dependencies~\cite{Yu2017}, (5) enlarging the receptive fields~\cite{Chen2018, Luo2016}, or subsampling, i.e. using pooling operations in exchange for losing resolution. The latter could be mitigated by using residual connections~\cite{He2016,Villegas2017a} to preserve resolution while increasing the number of stacking convolutions. But even addressing these limitations, would \glspl{cnn} be able to predict in a longer time horizon?

Vanilla~\glspl{cnn} lack of explicit inter-frame modeling capabilities. To properly model inter-frame variability in a video sequence, 3D convolutions come into play as a promising alternative to recurrent modeling. Several video prediction approaches leveraged 3D convolutions to capture temporal consistency~\cite{Wang2019b,Vondrick2016,Tulyakov2018,Aigner2018,Zhang2019}. Also modeling time dimension, Amersfoort \textit{et al.} \cite{Amersfoort2017} replicated a purely convolutional approach in time to address multi-scale predictions in the transformation space. In this case, the learned affine transforms at each time step play the role of a recurrent state.

\subsection{Recurrent Models}
Recurrent models were specifically designed to model a spatio-temporal representation of sequential data such as video sequences. Among other sequence learning tasks, such as machine translation, speech recognition and video captioning, to name a few, \glspl{rnn}~\cite{Rumelhart1986} demonstrated great success in the video prediction scenario~\cite{Ranzato2014,Srivastava2015,Shi2015,Lotter2017,Lotter2015,Byeon2018, Zhang2019,Patraucean2015,Lu2017,Villegas2017,Denton2017,Chen2017,Oh2015,Denton2018,Villegas2017,Nabavi2018,Vora2018,Sun2019,Terwilliger2019,Wichers2018,Minderer2019}. Vanilla~\glspl{rnn} have some important limitations when dealing with long-term representations due to the vanishing and exploding gradient issues, making the \gls{bptt} cumbersome. By extending classical \glspl{rnn} to more sophisticated recurrent models, such as \gls{lstm}~\cite{Hochreiter1997} and \gls{gru}~\cite{Cho2014}, these problems were mitigated. Shi \textit{et al.} extended the use of \gls{lstm}-based models to the image space~\cite{Shi2015}. While some authors explored \gls{mdlstm}~\cite{Graves2007}, others stacked recurrent layers to capture abstract spatio-temporal correlations~\cite{Lotter2015,Finn2016}. Zhang \textit{et al.} addressed the duplicated representations along the same recurrent paths~\cite{Zhan2019}.

\subsection{Generative Models}
\label{subsec:generative}
Whilst discriminative models learn the decision boundaries between classes, generative models learn the underlying distribution of individual classes. More formally, discriminative models capture the conditional probability $p(y|x)$, while generative models capture the joint probability $p(x,y)$, or $p(x)$ in the absence of labels $y$. The goal of generative models is the following: given some training data, generate new samples from the same distribution. Let input data~$\sim p_{data}(x)$ and generated samples~$\sim p_{model}(x)$ where, $p_{data}$ and $p_{model}$ are the underlying input data and model's probability distribution respectively. The training process consists in learning a $p_{model}(x)$ similar to $p_{data}(x)$. This is done by explicitly, e.g \acrshort{vae}s, or implicitly, e.g. \glspl{gan}, estimating a density function from the input data. In the context of video prediction, generative models are mainly used to cope with future uncertainty by generating a wide spectrum of feasible predictions rather than a single eventual outcome.

\subsubsection{Explicit Density Modeling}
These models explicitly define and solve for $p_{model}(x)$. 

\vspace*{0.15cm}\noindent\textbf{PixelRNNs and PixelCNNs~\cite{Oord2016}}: These are a type of \glspl{fvbn}~\cite{Neal1992,Bengio1999} that explicitly define a tractable density and estimate the joint distribution $p(x)$ as a product of conditional distributions over the pixels. Informally, they turn pixel generation into a sequential modeling problem, where next pixel values are determined by previously generated ones. In PixelRNNs, this conditional dependency on previous pixels is modeled using \gls{2d} \glspl{lstm}. On the other hand, dependencies are modeled using convolutional operations over a context region, thus making training faster. In a nutshell, these methods are outputting a distribution over pixel values at each location in the image, aiming to maximize the likelihood of the training data being generated. Further improvements of the original architectures have been carried out to address different issues. The Gated PixelCNN~\cite{Oord2016a} is computationally more efficient and improves the receptive fields of the original architecture. In the same work, authors also explored conditional modeling of natural images, where the joint probability distribution is conditioned on a latent vector ---it represents a high-level image description. This further enabled the extension to video prediction~\cite{Kalchbrenner2016}. 

\vspace*{0.15cm}\noindent\textbf{\glspl{vae}}: These models are an extension of \Glspl{ae} that encode and reconstruct its own input data $x$ in order to capture a low-dimensional representation $z$ containing the most meaningful factors of variation in $x$. Extending this architecture to generation, \glspl{vae} aim to sample new images from a prior over the underlying latent representation $z$. \glspl{vae} represent a probabilistic spin over the deterministic latent space in \glspl{ae}. Instead of directly optimizing the density function, which is intractable, they derive and optimize a lower bound on the likelihood. Data is generated from the learned distribution by perturbing the latent variables. In the video prediction context, \glspl{vae} are the foundation of many probabilistic models dealing with future uncertainty~\cite{Liang2017,Babaeizadeh2018,Denton2018,Fragkiadaki2017,Castrejon2019,Bhattacharyya2019,Minderer2019}. Although these variational approaches are able to generate various plausible outcomes, the predictions are blurrier and of lower quality compared to state-of-the-art \gls{gan}-based models. Many approaches were taken to leverage the advantages of variational inference: combined adversarial training with \glspl{vae}~\cite{Liang2017}, and others incorporated latent probabilistic variables into deterministic models, such as \glspl{vrnn}~\cite{Chung2015,Castrejon2019} and \glspl{ved}~\cite{Henaff2017}.

\subsubsection{Implicit Density Modeling}
These models learn to sample from $p_{model}$ without explicitly defining it. 

\vspace*{0.15cm}\noindent\textbf{\glspl{gan}}~\cite{Goodfellow2014}: are the backbone of many video prediction approaches~\cite{Mathieu2016,Lotter2015,Liang2017,Hu2019,Chen2017,Wichers2018,Jin2017,Walker2017,Villegas2017,Kwon2019,Vondrick2017, Villegas2017a,Lu2017,Zhou2016,Bhattacharjee2017,Vondrick2016,Saito2017,Chen2017a,Tulyakov2018}. Inspired on game theory, these networks consist of two models that are jointly trained as a minimax game to generate new fake samples that resemble the real data. On one hand, we have the discriminator model featuring a probability distribution function describing the real data. On the other hand, we have the generator which tries to generate new samples that fool the discriminator. In their original formulation, \glspl{gan} are unconditioned --the generator samples new data from a random noise, e.g. Gaussian noise. Nevertheless, Mirza \textit{et al.}~\cite{Mirza2014} proposed the \gls{cgan}, a conditional version where the generator and discriminator are conditioned on some extra information, e.g. class labels, previous predictions, and multimodal data, among others. \Glspl{cgan} are suitable for video prediction, since the spatio-temporal coherence between the generated frames and the input sequence is guaranteed. The use of adversarial training for the video prediction task, represented a leap over the previous state-of-the-art methods in terms of prediction quality and sharpness. However, adversarial training is unstable. Without an explicit latent variable interpretation, \glspl{gan} are prone to mode collapse ---generator fails to cover the space of possible predictions by getting stuck into a single mode~\cite{Henaff2017,Lee2018}. Moreover, \glspl{gan} often struggle to balance between the adversarial and reconstruction loss, thus getting blurry predictions. Among the dense literature on adversarial networks, we find some other interesting works addressing \glspl{gan} limitations~\cite{Radford2016,Arjovsky2017}. 
\section{Datasets}
\label{sec:datasets}
As video prediction models are mostly self-supervised, they need video sequences as input data. However, some video prediction methods rely on extra supervisory signals, e.g. segmentation maps, and human poses. This makes out-of-domain video datasets perfectly suitable for video prediction. This section describes the most relevant datasets, discussing their pros and cons. Datasets were organized according to their main purpose and summarized in Table~\ref{table:datasets}.

\subsection{Action and Human Pose Recognition Datasets}
\textbf{KTH~\cite{Schuldt2004}}: is an action recognition dataset which includes \num{2391} video sequences of \num{4} seconds mean duration, each of them containing an actor performing an action taken with a static camera, over homogeneous backgrounds, at \num{25} \gls{fps} and with its resolution downsampled to $160 \times 120$ pixels. Just \num{6} different actions are performed, but it was the biggest dataset of this kind at its moment.

\vspace*{0.15cm}\noindent\textbf{Weizmann~\cite{Gorelick2007}}: is also an action recognition dataset, created for modelling actions as space-time shapes. For this reason, it was recorded at a higher frame rate (\num{50} \gls{fps}). It just includes \num{90} video sequences, but performing 10 different actions. It uses a static-camera, homogeneous backgrounds and low resolution ($180 \times 144$ pixels). KTH and Weizmann are usually used together due to their similarities in order to augment the amount of available data.

\vspace*{0.15cm}\noindent\textbf{HMDB-51~\cite{Kuehne2011}}: is a large-scale database for human motion recognition. It claims to represent the richness of human motion taking profit from the huge amount of video available online. It is composed by \num{6766} normalized videos (with mean duration of \num{3.15} seconds) where humans appear performing one of the 51 considered action categories. Moreover, a stabilized dataset version is provided, in which camera movement is disabled by detecting static backgrounds and displacing the action as a window. It also provides interesting data for each sequence such as body parts visible, point of view respect the human, and if there is camera movement or not. It also exists a joint-annotated version called J-HMBD~\cite{Jhuang2013} in which the key points of joints were mannually added for 21 of the HMDB actions.

\vspace*{0.15cm}\noindent\textbf{UCF101~\cite{Soomro2012}}: is an action recognition dataset of realistic action videos, collected from YouTube. It has 101 different action categories, and it is an extension of UCF50, which has \num{50} action categories. All videos have a frame rate of 25 \gls{fps} and a resolution of $320 \times 240$ pixels. Despite being the most used dataset among predictive models, a problem it has is that only a few sequences really represent movement, i.e. they often show an action over a fixed background.

\vspace*{0.15cm}\noindent\textbf{Penn Action Dataset~\cite{Zhang2013}}: is an action and human pose recognition dataset from the University of Pennsylvania. It contains \num{2326} video sequences of \num{15} different actions, and it also provides human joint and viewpoint (position of the camera respect the human) annotations for each sequence. Each action is balanced in terms of different viewpoints representation.

\vspace*{0.15cm}\noindent\textbf{Human3.6M~\cite{Ionescu2014}}: is a human pose dataset in which \num{11} actors with marker-based suits were recorded performing \num{15} different types of actions. It features RGB images, depth maps (time-of-flight range data), poses and scanned 3D surface meshes of all actors. Silhouette masks and 2D bounding boxes are also provided. Moreover, the dataset was extended by inserting high-quality 3D rigged human models (animated with the previously recorded actions) in real videos, to create a realistic and complex background.

\begin{table*}[t] \centering
	\ra{1.15}
	\caption{Summary of the most widely used datasets for video prediction (\textbf{S/R}:~\textbf{S}ynthetic/\textbf{R}eal, \textbf{st}:~\textbf{st}ereo, \textbf{de}:~\textbf{de}pth,
		\textbf{ss}:~\textbf{s}emantic \textbf{s}egmentation, \textbf{is}:~\textbf{i}nstance \textbf{s}egmentation, \textbf{sem}:~\textbf{sem}antic, \textbf{I/O}:~\textbf{I}ndoor/\textbf{O}utdoor environment, \textbf{bb}:~\textbf{b}ounding \textbf{b}ox, \textbf{Act}:~\textbf{Act}ion label, \textbf{ann}:~\textbf{ann}otated, \textbf{env}:~\textbf{env}ironment, \textbf{ToF}:~\textbf{T}ime \textbf{o}f \textbf{F}light, \textbf{vp}:~camera \textbf{v}iew\textbf{p}oints respect human).}
	\label{table:datasets}
	\footnotesize
	\resizebox{\textwidth}{!}{
	\begin{threeparttable}
		\begin{tabular} {@{}lcccccccccccccc@{}} 
			\toprule
			& & & & & & & & \multicolumn{6}{c}{\textbf{provided data and ground-truth}} & \\
			\cmidrule(lr){9-14}
			\textbf{name}\tnote{1} & \textbf{year} & \textbf{S/R} & \textbf{\#videos} & \textbf{\#frames} & \textbf{\#ann. frames} & \textbf{resolution} & \textbf{\#classes} & \textbf{RGB} & \textbf{st} & \textbf{de} & \textbf{ss} & \textbf{is} & \textbf{other annotations} & \textbf{env.}\\
			\midrule
			\multicolumn{15}{c}{\textbf{Action and human pose recognition datasets}} \\
			\midrule
			KTH \cite{Schuldt2004} & \num{2004} & R & \num{2391} & \num{250000}\tnote{2} & \num{0} & $160 \times 120$ & \num{6} (action) & \checkmark & \xmark & \xmark & \xmark & \xmark & Act. & O \\
			Weizmann \cite{Gorelick2007} & \num{2007} & R & \num{90} & \num{9000}\tnote{2} & \num{0} & $180 \times 144$ & \num{10} (action) & \checkmark & \xmark & \xmark & \xmark & \xmark & Act. & O  \\
			HMDB-51 \cite{Kuehne2011} & \num{2011} & R & \num{6766} & \num{639300} & \num{0} & $var~\times 240$ & \num{51} (action) & \checkmark & \xmark & \xmark & \xmark & \xmark & Act., vp & I/O\\
			UCF101 \cite{Soomro2012} & \num{2012} & R & \num{13320} & \num{2000000}\tnote{2} & \num{0} & $320 \times 240$ & \num{101} (action) & \checkmark & \xmark & \xmark & \xmark & \xmark & Act. & I/O \\
			Penn Action D. \cite{Zhang2013} & \num{2013} & R & \num{2326} & \num{163841} & \num{0} & $480 \times 270$ & \num{15} (action) & \checkmark & \xmark & \xmark & \xmark & \xmark & Act., Human poses, vp & I/O \\
			Human3.6M \cite{Ionescu2014} & \num{2014} & SR & \num{4000}\tnote{2} & \num{3600000} & \num{0} & $1000 x 1000$ & \num{15} (action) & \checkmark & \xmark & ToF & \xmark & \xmark & Act., Human poses \& meshes\comment{Skeleton poses, Human meshes} & I/O  \\
			THUMOS-15 \cite{Idrees2017} & \num{2017} & R & \num{18404} & \num{3000000}\tnote{2} & \num{0} & $320 \times 240$ & \num{101} (action) & \checkmark & \xmark & \xmark & \xmark & \xmark & Act., Time span & I/O \\
			\midrule
			\multicolumn{15}{c}{\textbf{Driving and urban scene understanding datasets}} \\
			\midrule
			Camvid \cite{Patraucean2015} & \num{2008} & R & \num{5} & \num{18202} & \num{701} (ss) & $960 \times 720$ & \num{32} (sem) & \checkmark & \xmark & \xmark & \checkmark & \xmark & \xmark & O \\
			CalTech Pedest. \cite{Dollar2009} & \num{2009} & R & \num{137} & \num{1000000}\tnote{2} & \num{250000} (bb) & $640 \times 480$ & - & \checkmark & \xmark & \xmark & \xmark & \xmark & Pedestrian bb \& occlusions & O \\
			Kitti \cite{Geiger2013} & \num{2013} & R & \num{151} & \num{48791} & \num{200} (ss) & $1392 \times 512$ & \num{30} (sem) & \checkmark & \checkmark & LiDAR & \checkmark & \checkmark & Odometry & O  \\
			Cityscapes \cite{Cordts2016} & \num{2016} & R & \num{50} & \num{7000000}\tnote{2} & \num{25000} (ss) & $2048 \times 1024$ & \num{30} (sem) & \checkmark & \checkmark & stereo & \checkmark & \checkmark & Odometry, temp, GPS & O \\
			Comma.ai \cite{Lotter2017} & \num{2016} & R & \num{11} & \num{522000}\tnote{2} & \num{0} & $160 \times 320$ & - & \checkmark & \xmark & \xmark & \xmark & \xmark & Steering angles \& speed & O \\
			Apolloscape \cite{Huang2018a} & \num{2018} & R & \num{4} & \num{200000} & \num{146997} (ss) & $3384 \times 2710$ & \num{25} (sem) & \checkmark & \checkmark & LiDAR & \checkmark & \checkmark & Odometry, GPS & O \\	
			\midrule
			\multicolumn{15}{c}{\textbf{Object and video classification datasets}} \\
			\midrule
			Sports1m \cite{Karpathy2014} & \num{2014} & R & \num{1133158} & n/a & \num{0} & $640 \times 360$ $(var.)$ & \num{487} (sport) & \checkmark & \xmark & \xmark & \xmark & \xmark & Sport label & I/O \\
			YouTube8M \cite{Abu-El-Haija2016} & \num{2016} & R & \num{8200000} & n/a & \num{0} & $variable$ & \num{1000} (topic) & \checkmark & \xmark & \xmark & \xmark & \xmark & Topic label, Segment info & I/O \\
			YFCC100M \cite{Thomee2016} & \num{2016} & SR & \num{8000} & n/a & \num{0} & $variable$ & - & \checkmark & \xmark & \xmark & \xmark & \xmark  & User tags, Localization & I/O \\
			\midrule
			\multicolumn{15}{c}{\textbf{Video prediction datasets}} \\
			\midrule
			Bouncing balls \cite{Sutskever2008} & \num{2008} & S & \num{4000} & \num{20000} & \num{0} & $150 \times 150$ & - & \checkmark & \xmark & \xmark & \xmark & \xmark & \xmark & - \\
			Van Hateren \cite{Cadieu2012} & \num{2012} & R & \num{56} & \num{3584} & \num{0} & $128 \times 128$ & - & \checkmark & \xmark & \xmark & \xmark & \xmark & \xmark & I/O \\
			NORBvideos \cite{Memisevic2013} & \num{2013} & R & \num{110560} & \num{552800} & All (is) & $640 \times 480$ & \num{5} (object) & \checkmark & \xmark & \xmark & \xmark & \checkmark & \xmark & I \\
			Moving MNIST \cite{Srivastava2015} & \num{2015} & SR & $custom$\tnote{3} & $custom$\tnote{3} & \num{0} & $64 \times 64$ & - & \checkmark & \xmark & \xmark & \xmark & \xmark & \xmark & - \\
			Robotic Pushing \cite{Finn2016} & \num{2016} & R & \num{57000} & \num{1500000}\tnote{2} & \num{0} & $640 \times 512$ & - & \checkmark & \xmark & \xmark & \xmark & \xmark & Arm pose & I \\
			BAIR Robot \cite{Ebert2017} & \num{2017} & R & \num{45000} & n/a & \num{0} & n/a & - & \checkmark & \xmark & \xmark & \xmark & \xmark &  Arm pose & I \\
			RoboNet \cite{Dasari2019} & \num{2019} & R & \num{161000} & \num{15000000} & \num{0} & $variable$ & - & \checkmark & \xmark & \xmark & \xmark & \xmark & Arm pose & I \\
			\midrule
			\multicolumn{15}{c}{\textbf{Other-purpose and multi-purpose datasets}} \\
			\midrule
			ViSOR \cite{Vezzani2010} & \num{2010} & R & \num{1529} & \num{1360000}\tnote{2} & \num{0} & $variable$ & - & \checkmark & \xmark & \xmark & \xmark & \xmark & User tags, human bb & I/O \\
			PROST \cite{Santner2010} & \num{2010} & R & \num{4} (\num{10}) & \num{4936} (\num{9296}) & All (bb) & $variable$ & - & \checkmark & \xmark & \xmark & \xmark & \xmark & Object bb & I \\
			Arcade Learning \cite{Bellemare2013} & \num{2013} & S & $custom$\tnote{3} & $custom$\tnote{3} & \num{0} & $210 \times 160$ & - & \checkmark & \xmark & \xmark & \xmark & \xmark & \xmark & - \\
			Inria 3DMovie v2 \cite{Seguin2016} & \num{2016} & R & \num{27} & \num{2476} & \num{235} (is) & $960 \times 540$ & - & \checkmark & \checkmark & \xmark & \xmark & \checkmark & Human poses, bb & I/O \\
			Robotrix \cite{Garcia2018} & \num{2018} & S & \num{67} & \num{3039252} & All (ss) & $1920 \times 1080$ & \num{39} (sem) & \checkmark & \xmark & \checkmark & \checkmark & \checkmark & Normal maps, 6D poses & I \\
			UASOL \cite{Bauer2019} & \num{2019} & R & \num{33} & \num{165365} & \num{0} & $2280 \times 1282$ & - & \checkmark & \checkmark & stereo & \xmark & \xmark & \xmark & O \\
			\bottomrule
	\end{tabular}
	\begin{tablenotes}
	\item[1] some dataset names have been abbreviated to enhance table's readability.
	
	\item[2] values estimated based on the framerate and the total number of frames or videos, as the original values are not provided by the authors.
    
    \item[3] \textit{custom} indicates that as many frames as needed can be generated. This is related to datasets generated from a game, algorithm or simulation, involving interaction or randomness.
    \end{tablenotes}
    \end{threeparttable}}
\end{table*}

\vspace*{0.15cm}\noindent\textbf{THUMOS-15~\cite{Idrees2017}}: is an action recognition challenge that was celebrated in 2015. It didn't just focus on recognizing an action in a video, but also on determining the time span in which that action occurs. With that purpose, the challenge provided a dataset that extends UCF101~\cite{Soomro2012} (trimmed videos with one action) with 2100 untrimmed videos where one or more actions take place (with the correspondent temporal annotations) and almost \num{3000} relevant videos without any of the \num{101} proposed actions.

\subsection{Driving and Urban Scene Understanding Datasets}
\textbf{CamVid~\cite{Brostow2008}}: the Cambridge-driving Labeled Video Database is a driving/urban scene understanding dataset which consists of \num{5} video sequences recorded with a $960 \times 720$ pixels resolution camera mounted on the dashboard of a car. Four of those sequences were sampled at \num{1} \gls{fps}, and one at \num{15} \gls{fps}, resulting in \num{701} frames which were manually per-pixel annotated for semantic segmentation (under \num{32} classes). It was the first video sequence dataset of this kind to incorporate semantic annotations.

\vspace*{0.15cm}\noindent\textbf{CalTech Pedestrian Dataset~\cite{Dollar2009}}: is a driving dataset focused on detecting pedestrians, since its unique annotations are pedestrian bounding boxes. It is conformed of approximately \num{10} hours of $640 \times 480$ $30$\gls{fps} video taken from a vehicle driving through regular traffic in an urban environment, making a total of \num{250000} annotated frames distributed in \num{137} approximately minute-long segments. The total pedestrian bounding boxes is \num{350000}, identifying \num{2300} unique pedestrians. Temporal correspondence between bounding boxes and detailed occlusion labels are also provided.

\vspace*{0.15cm}\noindent\textbf{Kitti~\cite{Geiger2013}}: is one of the most popular datasets for mobile robotics and autonomous driving, as well as a benchmark for computer vision algorithms. It is composed by hours of traffic scenarios recorded with a variety of sensor modalities, including high-resolution RGB, gray-scale stereo cameras, and a 3D laser scanner. Despite its popularity, the original dataset did not contain ground truth for semantic segmentation. However, after various researchers manually annotated parts of the dataset to fit their necessities, in 2015 Kitti dataset was updated with 200 annotated frames at pixel level for both semantic and instance segmentation, following the format proposed by the Cityscapes~\cite{Cordts2016} dataset.

\vspace*{0.15cm}\noindent\textbf{Cityscapes~\cite{Cordts2016}}: is a large-scale database which focuses on semantic understanding of urban street scenes. It provides semantic, instance-wise, and dense pixel annotations for \num{30} classes grouped into \num{8} categories. The dataset consist of around \num{5000} fine annotated images (\num{1} frame in \num{30}) and \num{20000} coarse annotated ones (one frame every \num{20} seconds or \num{20} meters run by the car). Data was captured in \num{50} cities during several months, daytimes, and good weather conditions. All frames are provided as stereo pairs, and the dataset also includes vehicle odometry obtained from in-vehicle sensors, outside temperature, and GPS tracks.

\vspace*{0.15cm}\noindent\textbf{Comma.ai steering angle~\cite{Santana2016}}: is a driving dataset composed by \num{7.25} hours of largely highway routes. It was recorded as $360 \times 180$ camera images at 20 \gls{fps} (divided in \num{11} different clips), and steering angles, among other vehicle data (speed, GPS, etc.).

\vspace*{0.15cm}\noindent\textbf{Apolloscape~\cite{Huang2018a}}: is a driving/urban scene understanding dataset that focuses on 3D semantic reconstruction of the environment. It provides highly precise information about location and 6D camera pose, as well as a much bigger amount of dense per-pixel annotations than other datasets. Along with that, depth information is retireved from a LIDAR sensor, that allows to semantically reconstruct the scene in 3D as a point cloud. It also provides RGB stereo pairs as video sequences recorded under various weather conditions and daytimes. This video sequences and their per-pixel instance annotations make this dataset very interesting for a wide variety of predictive models.

\vspace*{0.15cm}\noindent\subsection{Object and Video Classification Datasets}
\textbf{Sports1M~\cite{Karpathy2014}}: is a video classification dataset that also consists of annotated YouTube videos. In this case, it is fully focused on sports: its \num{487} classes correspond to the sport label retrieved from the YouTube Topics API. Video resolution, duration and frame rate differ across all available videos, but they can be normalized when accessed from YouTube. It is much bigger than UCF101 (over 1 million videos), and movement is also much more frequent.

\vspace*{0.15cm}\noindent\textbf{Youtube-8M~\cite{Abu-El-Haija2016}}: Sports1M~\cite{Karpathy2014} dataset is, since 2016, part of a bigger one called YouTube8M, which follows the same philosophy, but with all kind of videos, not just sports. Moreover, it has been updated in order to improve the quality and precision of their annotations. In 2019 YouTube-8M Segments was released with segment-level human-verified labels on about \num{237000} video segments on \num{1000} different classes, which are collected from the validation set of the YouTube-8M dataset. Since YouTube is the biggest video source on the planet, having annotations for some of their videos at segment level is great for predictive models.

\vspace*{0.15cm}\noindent\textbf{YFCC100M~\cite{Thomee2016}}: \textit{Yahoo Flickr Creative Commons 100 Million Dataset} is a collection of \num{100} million images and videos uploaded to Flickr between 2004 and 2014. All those media files were published in  Flickr under Creative Commons license, overcoming one of the biggest issues affecting existing multimedia datasets, licensing and volume. Although only 0.8\% of the elements of the dataset are videos, it is still useful for predictive models due to the great variety of these, and therefore the challenge that it represents.

\subsection{Video Prediction Datasets}
\textbf{Standard bouncing balls dataset~\cite{Sutskever2008}}: is a common test set for models that generate high dimensional sequences. It consists of simulations of three balls bouncing in a box. Its clips can be generated randomly with custom resolution but the common structure is composed by \num{4000} training videos, \num{200} testing videos and \num{200} more for validation. This kind of datasets are purely focused on video prediction.

\vspace*{0.15cm}\noindent\textbf{Van Hateren Dataset of natural videos (version~\cite{Cadieu2012})}: is a very small dataset of \num{56} videos, each \num{64} frames long, that has been widely used in unsupervised learning. Original images were taken and given for scientific use by the photographer Hans van Hateren, and they feature moving animals in grasslands along rivers and streams. Its frame size is $128 \times 128$ pixels. The version we are reviewing is the one provided along with the work of Cadieu and Olshausen~\cite{Cadieu2012}.

\vspace*{0.15cm}\noindent\textbf{NORBvideos~\cite{Memisevic2013}}: NORB (NYU Object Recognition Benchmark) dataset~\cite{LeCun2004} is a compilation of static stereo pairs of 50 homogeneously colored objects from various points of view and 6 lightning conditions. Those images were processed to obtain their object masks and even their casted shadows, allowing them to augment the dataset introducing random backgrounds. Viewpoints are determined by rotating the camera through 9 elevations and 18 azimuths (every 20 degrees) around the object. \textit{NORBvideos} dataset was built by sequencing all these frames for each object.

\vspace*{0.15cm}\noindent\textbf{Moving MNIST~\cite{Srivastava2015} (M-MNIST)}: is a video prediction dataset built from the composition of 20-frame video sequences where two handwritten digits from the MNIST database are combined inside a $64 \times 64$ patch, and moved with some velocity and direction along frames, potentially overlapping between them. This dataset is almost infinite (as new sequences can be generated on the fly), and it also has interesting behaviours due to occlusions and the dynamics of digits bouncing off the walls of the patch. For these reasons, this dataset is widely used by many predictive models. A stochastic variant of this dataset is also available. In the original M-MNIST the digits move with constant velocity and bounce off the walls in a deterministic manner. In contrast, in SM-MNIST digits move with a constant velocity along a trajectory until they hit at wall at which point they bounce off with a random speed and direction. In this way, moments of uncertainty (each time a digit hits a wall) are interspersed with deterministic motion.

\vspace*{0.15cm}\noindent\textbf{Robotic Pushing Dataset~\cite{Finn2016}}: is a dataset created for learning about physical object motion. It consist on $640 \times 512$ pixels image sequences of \num{10} different 7-degree-of-freedom robotic arms interacting with real-world physical objects. No additional labeling is given, the dataset was designed to model motion at pixel level through deep learning algorithms based on \gls{convlstm}.

\vspace*{0.15cm}\noindent\textbf{BAIR Robot Pushing Dataset (used in \cite{Ebert2017})}: BAIR (Berkeley Artificial Intelligence Research) group has been working on robots that can learn through unsupervised training (also known in this case as self-supervised), this is, learning the consequences that its actions (movement of the arm and grip) have over the data it can measure (images from two cameras). In this way, the robot assimilates physics of the objects and can predict the effects that its actions will generate on the environment, allowing it to plan strategies to achieve more general goals. This was improved by showing the robot how it can grab tools to interact with other objects. The dataset is composed by hours of this self-supervised learning with the robotic arm \textit{Sawyer}.

\vspace*{0.15cm}\noindent\textbf{RoboNet~\cite{Dasari2019}}: is a dataset composed by the aggregation of various self-supervised training sequences of seven robotic arms from four different research laboratories. The previously described BAIR group is one of them, along with \textit{Stanford AI Laboratory, \textit{Grasp Lab of the University of Pennsylvania} and \textit{Google Brain Robotics}}. It was created with the goal of being a standard, in the same way as ImageNet is for images, but for robotic self-supervised learning. Several experiments have been performed studying how the transfer among robotic arms can be achieved.

\vspace*{0.15cm}\noindent\subsection{Other-purpose and Multi-purpose Datasets}
\textbf{ViSOR~\cite{Vezzani2010}}: ViSOR (Video Surveillance Online Repository) is a repository designed with the aim of establishing an open platform for collecting, annotating, retrieving, and sharing surveillance videos, as well as evaluating the performance of automatic surveillance systems. Its raw data could be very useful for video prediction due to its implicit static camera.

\vspace*{0.15cm}\noindent\textbf{PROST~\cite{Santner2010}}: is a method for online tracking that used ten manually annotated videos to test its performance. Four of them were created by PROST authors, and they conform the dataset with the same name. The remaining six sequences were borrowed from other authors, who released their annotated clips to test their tracking methods. We will consider both 4-sequences PROST dataset and 10-sequences aggregated dataset when providing statistics. In each video, different challenges are presented for tracking methods: occlusions, 3D motion, varying illumination, heavy appearance/scale changes, moving camera, motion blur, among others. Provided annotations include bounding boxes for the object/element being tracked.

\vspace*{0.15cm}\noindent\textbf{Arcade Learning Environment \cite{Bellemare2013}}: is a platform that enables machine learning algorithms to interact with the Atari \num{2600} open-source emulator Stella to play over \num{500} Atari games. The interface provides a single 2D frame of $210 \times 160$ pixels resolution at \num{60} \gls{fps} in real-time, and up to \num{6000} \gls{fps} when it is running at full speed. It also offers the possibility of saving and restoring the state of a game. Although its obvious main application is reinforcement learning, it could also be profitable as source of almost-infinite interactive video sequences from which prediction models can learn.

\vspace*{0.15cm}\noindent\textbf{Inria 3DMovie Dataset v2~\cite{Seguin2016}}: is a video dataset which extracted its data from the \textit{StreetDance 3D} stereo movies. The dataset includes stereo pairs, and manually generated ground-truth for human segmentation, poses and bounding boxes. The second version of this dataset, used in~\cite{Seguin2016}, is composed by \num{27} clips, which represent \num{2476} frames, of which just a sparse subset of \num{235} were annotated.

\vspace*{0.15cm}\noindent\textbf{RobotriX~\cite{Garcia2018}}: is a synthetic dataset designed for assistance robotics, that consist of sequences where a humanoid robot is moving through various indoor scenes and interacting with objects, recorded from multiple points of view, including robot-mounted cameras. It provides a huge variety of ground-truth data generated synthetically from highly-realistic environments deployed on the cutting-edge game engine UnrealEngine, through the also available tool UnrealROX~\cite{Martinez2019}. RGB frames are provided at $1920 \times 1080$ pixels resolution and at \num{60} \gls{fps}, along with pixel-precise instance masks, depth and normal maps, and 6D poses of objects, skeletons and cameras. Moreover, UnrealROX is an open source tool for retrieving ground-truth data from any simulation running in UnrealEngine.

\vspace*{0.15cm}\noindent\textbf{UASOL~\cite{Bauer2019}}: is a large-scale dataset consisting of high-resolution sequences of stereo pairs recorded outdoors at pedestrian (egocentric) point of view. Along with them, precise depth maps are provided, computed offline from stereo pairs by the same camera. This dataset is intended to be useful for depth estimation, both from single and stereo images, research fields where outdoor and pedestrian-point-of-view data is not abundant. Frames were taken at a resolution of $2280 \times 1282$ pixels at \num{15} \gls{fps}.

\colorlet{linecol}{black!65}
\tikzset{
	my rounded corners/.append style={rounded corners=2pt},
}
\begin{figure*}[ht!]
\centering
\resizebox{0.9\linewidth}{!}{
\begin{forest}
		for tree={line width=1pt,
			if={level()<2}{my rounded corners, draw=linecol,}{},edge={color=linecol, >={Triangle[]}, ->},
			if level=0{%
				fill=gray!10,
				l sep+=0.15cm,
				s sep+=0.5cm,
				align=center,
				parent anchor=south}{%
				if level=1{%
					parent anchor=south west,
					child anchor=north,
					tier=parting ways,
					align=center,
					minimum width= 3.5cm,
					fill=gray!20,
					font=\bfseries,
					l sep-= 8pt,
					for descendants={
						child anchor=west,
						parent anchor=west,
						anchor=west,
						align=left,
						s sep-= 8pt,
					},
				}{
					if level=2{
						shape=coordinate,
						no edge,
						grow'=0,
						calign with current edge,
						xshift=25pt,
						for descendants={
							parent anchor=south west,
						},
						for children={
							edge path={
								\noexpand\path[\forestoption{edge}] (!to tier=parting ways.parent anchor) |- (.child anchor)\forestoption{edge label};
							},
							for descendants={
								no edge,
							},
						},
					}{},
				},
			}%
		},
	[Video Prediction
		[Through Direct\\ Pixel Synthesis
			[
				[Implicit Modeling\\ of Scene Dynamics]
			]
		]
		[Factorizing the\\ Prediction Space
			[
				[Using Explicit\\ Transformations]
				[With Explicit Mo-\\tion from Content\\ Separation]
			]
		]
		[Narrowing the\\ Prediction Space
			[
				[By Conditioning on\\ Extra Variables]
				[To High-level\\ Feature Space]
			]
		]
		[By Incorporating\\ Uncertainty
			[
				[Using Probabilistic\\ Approaches]
			]
		]
	]
\end{forest}}
\caption{Classification of video prediction models.}
\label{fig:classification}
\end{figure*}
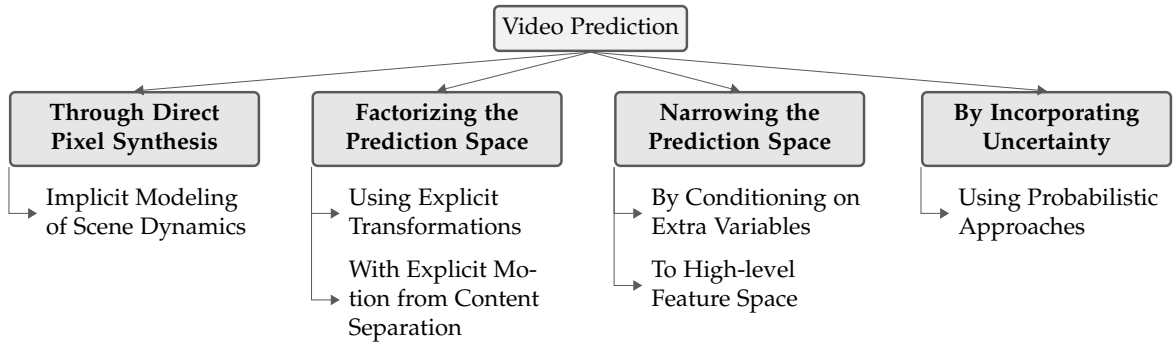

\section{Video Prediction Methods}
\label{sec:prediction_methods}

In the video prediction literature we find a broad range of different methods and approaches. Early models focused on directly predicting raw pixel intensities, by implicitly modeling scene dynamics and low-level details (Section~\ref{subsec:m_raw}). However, extracting a meaningful and robust representation from raw videos is challenging, since the pixel space is highly dimensional and extremely variable. From this point, reducing the supervision effort and the representation dimensionality emerged as a natural evolution. On the one hand, the authors aimed to disentangle the factors of variation from the visual content, i.e. factorizing the prediction space. For this purpose, they: (1) formulated the prediction problem into an intermediate transformation space by explicitly modeling the source of variability as transformations between frames (Section~\ref{subsec:m_transform}); (2) separated motion from the visual content with a two-stream computation (Section~\ref{subsc:two_stream}). On the other hand, some models narrowed the output space by conditioning the predictions on extra variables (Section~\ref{subsec:m_condition}), or reformulating the problem in a higher-level space (Section~\ref{subsec:m_high_level}). High-level representation spaces are increasingly more attractive, since intelligent systems rarely rely on raw pixel information for decision making. Besides simplifying the prediction task, some other works addressed the future uncertainty in predictions. As the vast majority of video prediction models are deterministic, they are unable to manage probabilistic environments. To address this issue, several authors proposed modeling future uncertainty with probabilistic models (Section~\ref{subsec:m_uncertainty}). 

So far in the literature, there is no specific taxonomy that classifies video prediction models. In this review, we have classified the existing methods according to the video prediction problem they addressed and following the classification illustrated in Figure~\ref{fig:classification}. For simplicity, each subsection extends directly the last level in the taxonomy. Moreover, some methods in this review can be classified in more than one category since they addressed multiple problems. For instance, \cite{Walker2017,Bhattacharyya2019,Minderer2019} are probabilistic models making predictions in a high-level space as they addressed both the future uncertainty and high dimensionality in videos. The category of these models were specified according to their main contribution. The most relevant methods, ordered in a chronological order, are summarized in Table~\ref{table:methods} containing low-level details.
Prediction is a widely discussed topic in different fields and at different levels of abstraction. For instance, the future prediction from a static image~\cite{Vondrick2016a,Jayaraman2016,Walker2016,Chen2017a,Hao2018,Ye2019}, vehicle behavior prediction~\cite{Mozaffari2019} and human action prediction~\cite{Kong2018} are a different but inspiring research fields. Although related, the aforementioned topics are outside the scope of this particular review, as it focuses purely on the video prediction methods using a sequence of previous frames as context and is limited to 2D RGB data.

\subsection{Direct Pixel Synthesis}
\label{subsec:m_raw}
Initial video prediction models attempted to directly predict future pixel intensities without any explicit modeling of the scene dynamics. Ranzato \textit{et al.}~\cite{Ranzato2014} discretized video frames in patch clusters using k-means. They assumed that non-overlapping patches are equally different in a k-means discretized space, yet similarities can be found between patches. The method is a convolutional extension of a \gls{rnn}-based model~\cite{Mikolov2010} making short-term predictions at the patch-level. As the full-resolution frame is a composition of the predicted patches, some tilling effect can be noticed. Predictions of large and fast-moving objects are accurate, however, when it comes to small and slow-moving objects there is still room for improvement. These are common issues for most methods making predictions at the patch-level. Addressing longer-term predictions, Srivastava \textit{et al.}~\cite{Srivastava2015} proposed different \gls{ae}-based approaches incorporating \gls{lstm} units to model the temporal coherence. Using convolutional~\cite{Simonyan2015} and flow~\cite{Brox2004} percepts alongside RGB image patches, authors tested the models on multi-domain tasks and considered both unconditioned and conditioned decoder versions. The latter only marginally improved the prediction accuracy. Replacing the fully connected \glspl{lstm} with convolutional \glspl{lstm}, Shi \textit{et al.} proposed an end-to-end model efficiently exploiting spatial correlations~\cite{Shi2015}. This enhanced prediction accuracy and reduced the number of parameters.

\vspace*{0.3cm}\noindent\textbf{Inspired on adversarial training}: Building on the recent success of the \gls{lapgan}, Mathieu \textit{et al.} proposed the first multi-scale architecture for video prediction that was trained in an adversarial fashion~\cite{Mathieu2016}. Their novel \gls{gdl} regularization combined with $\ell_1$-based reconstruction and adversarial training represented a leap over the previous state-of-the-art models~\cite{Ranzato2014,Srivastava2015} in terms of prediction sharpness. However, it was outperformed by the \gls{prednet}~\cite{Lotter2017} which stacked several \glspl{convlstm} vertically connected by a bottom-up propagation of the local $\ell_1$ error computed at each level. Previously to \gls{prednet}, the same authors proposed the \gls{pgn}~\cite{Lotter2015}, an end-to-end model trained with a weighted combination of adversarial loss and~\gls{mse} on synthetic data. However, no tests on natural videos and comparison with state-of-the-art predictive models were carried out. Using a similar training strategy as~\cite{Mathieu2016}, Zhou \textit{et al.} used a convolutional \gls{ae} to learn long-term dependencies from time-lapse videos~\cite{Zhou2016}. Build on \glspl{pggan}~\cite{Karras2018}, Aigner \textit{et al.} proposed the FutureGAN~\cite{Aigner2018}, a \gls{3d} convolutional \gls{ed}-based model. They used the \gls{wgangp} loss~\cite{Gulrajani2017} and conducted experiments on increasingly complex datasets. Extending~\cite{Shi2015}, Zhang~\textit{et al.} proposed a novel \gls{lstm}-based architecture where hidden states are updated along a z-order curve~\cite{Zhang2019}. Dealing with distortion and temporal inconsistency in predictions and inspired by the \gls{hvs}, Jin \textit{et al.}~\cite{Jin2020} first incorporated multi-frequency analysis into the video prediction task to decompose images into low and high frequency bands. This allowed high-fidelity and temporally consistent predictions with the ground truth, as the model better leverages the spatial and temporal details. The proposed method outperformed previous state-of-the-art in all metrics except in the \gls{lpips}, where probabilistic models achieved a better performance since their predictions are clearer and realistic but less consistent with the ground truth. Distortion and blurriness are further accentuated when it comes to predict under fast camera motions. To this end, Shouno~\cite{Shouno2020} implemented a hierarchical residual network with top-down connections. Leveraging parallel prediction at multiple scales, authors reported finer details and textures under fast and large camera motion.

\vspace*{0.3cm}\noindent\textbf{Bidirectional flow}: Under the assumption that video sequences are symmetric in time, Kwon \textit{et al.}~\cite{Kwon2019} explored a retrospective prediction scheme training a generator for both, forward and backward prediction (reversing the input sequence to predict the past). Their cycle \gls{gan}-based approach ensure the consistency of bidirectional prediction through retrospective cycle constraints. Similarly, Hu \textit{et al.}~\cite{Hu2019} proposed a novel cycle-consistency loss used to train a \gls{gan}-based approach (VPGAN). Future frames are generated from a sequence of context frames and their variation in time, denoted as $Z$. Under the assumption that $Z$ is symmetric in the encoding space, it is manipulated by the model manipulates to generate desirable moving directions. In the same spirit, other works focused on both, forward and backward predictions~\cite{Misra2016,Hou2019}. Enabling state sharing between the encoder and decoder, Oliu \textit{et al.} proposed the \gls{frnn}~\cite{Oliu2018}, a recurrent \gls{ae} architecture featuring \glspl{gru} that implement a bidirectional flow of the information. The model demonstrated a stratified representation, which makes the topology more explainable, as well as efficient compared to regular \glspl{ae} in terms or memory consumption and computational requirements.

\begin{figure}[tbp]
	\centering
	\includegraphics[width=0.75\linewidth]{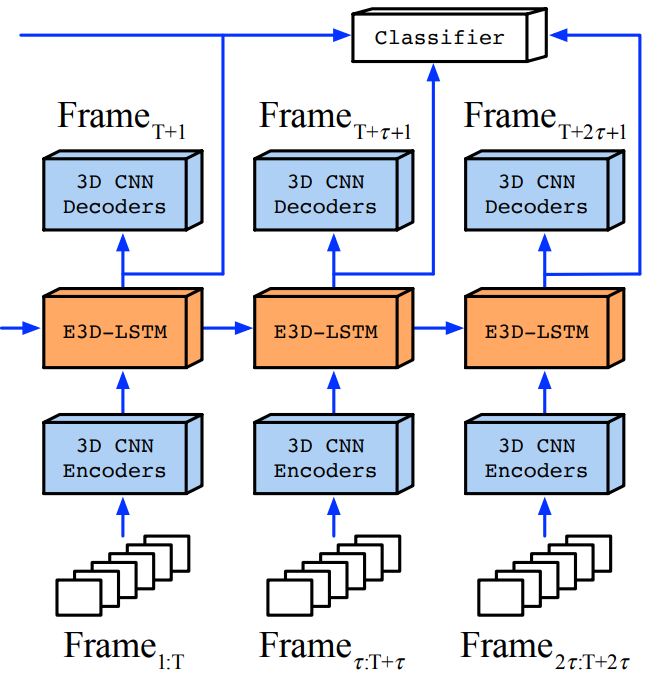}
	\caption{Representation of the 3D encoder-decoder architecture of \gls{e3dlstm} \cite{Wang2019b}. After reducing $T$ consecutive input frames to high-dimensional feature maps, these are directly fed into a novel \textit{eidetic} module for modeling long-term spatiotemporal dependencies. Finally, stacked 3D \gls{cnn} decoder outputs the predicted video frames. For classification tasks the hidden states can be directly used as the learned video representation. Figure extracted from~\cite{Wang2019b}.}
	\label{fig:e3d_lstm}
\end{figure}

\vspace*{0.3cm}\noindent\textbf{Exploiting 3D convolutions}: for modeling short-term features, Wang \textit{et al.}~\cite{Wang2019b} integrated them into a recurrent network demonstrating promising results in both video prediction and early activity recognition. While 3D convolutions efficiently preserves local dynamics, \glspl{rnn} enables long-range video reasoning. The \gls{e3dlstm} network, represented in Figure~\ref{fig:e3d_lstm}, features a gated-controlled self-attention module, i.e. \textit{eidetic} 3D memory, that effectively manages historical memory records across multiple time steps. Outperforming previous works, Yu~\textit{et al.} proposed the \gls{crevnet}~\cite{Yu2020} consisting of two modules, an invertible \gls{ae} and a \gls{rpm}. While the bijective two-way \gls{ae} ensures no information loss and reduces the memory consumption, the \gls{rpm} extends the reversibility from spatial to temporal domain. Some other works used 3D convolutional operations to model the time dimension~\cite{Aigner2018}.

Analyzing the previous works, Byeon \textit{et al.}~\cite{Byeon2018} identified a lack of spatial-temporal context in the representations, leading to blurry results when it comes to the future uncertainty. Although authors addressed this contextual limitation with dilated convolutions and multi-scale architectures, the context representation progressively vanishes in long-term predictions. To address this issue, they proposed a context-aware model that efficiently aggregates per-pixel contextual information at each layer and in multiple directions. The core of their proposal is a context-aware layer consisting of two blocks, one aggregating the information from multiple directions and the other blending them into a unified context.

Extracting a robust representation from raw pixel values is an overly complicated task due to the high-dimensionality of the pixel space. The per-pixel variability between consecutive frames, causes an exponential growth in the prediction error on the long-term horizon.

\subsection{Using Explicit Transformations}
\label{subsec:m_transform}
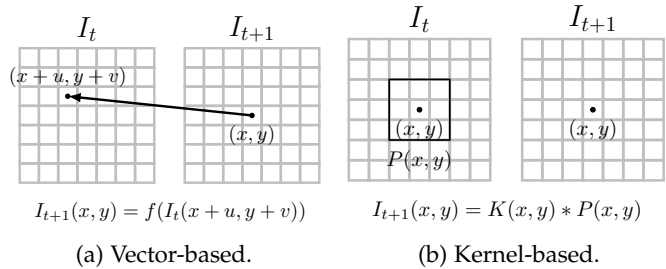
\begin{figure}[tbp]
	\centering
	\begin{subfigure}[t]{0.5\linewidth}
		\resizebox{\linewidth}{!}{
			\begin{tikzpicture}[]
	\tikzset{box/.style={black, draw=black, fill=none, very thick, minimum height=4 em, minimum width=16 em}}

	\foreach \k in {0,...,1}
	{
		\foreach \i in {0,...,6}
		{
			\foreach \j in {0,...,6}
			{
				\node[box, fill=none, draw=gray!50, minimum height=1 em, minimum width=1 em] (l1-\i-\j-\k) at (\i em + 0.9 em * \k em, \j em) {};
			}
		}
	}

	\node at ([yshift=1.5 em]l1-3-6-0) {\Large $I_t$};
	\node at ([yshift=1.5 em]l1-3-6-1) {\Large $I_{t+1}$};
	\node at ([yshift=-2 em, xshift=-1.25 em]l1-0-0-1) {$I_{t+1}(x, y) = f(I_t(x + u, y + v))$};
	\node[shape=circle, inner sep=0.1 em, fill=black] at (l1-2-4-0) {};
	\node[shape=circle, inner sep=0.1 em, fill=black] at (l1-3-3-1) {};

	\draw[-latex, very thick] (l1-3-3-1.center) node[yshift=-1 em]{$(x,y)$} -- (l1-2-4-0.center) node[yshift=1 em]{$(x+u, y+v)$};







\end{tikzpicture}
		}
		\caption{Vector-based.}
		\label{subfig:resampling_vector}
	\end{subfigure}~
	\begin{subfigure}[t]{0.5\linewidth}
		\resizebox{\linewidth}{!}{
			\begin{tikzpicture}[]
	\tikzset{box/.style={black, draw=black, fill=none, very thick, minimum height=4 em, minimum width=16 em}}

	\foreach \k in {0,...,1}
	{
		\foreach \i in {0,...,6}
		{
			\foreach \j in {0,...,6}
			{
				\node[box, fill=none, draw=gray!50, minimum height=1 em, minimum width=1 em] (l1-\i-\j-\k) at (\i em + 0.9 em * \k em, \j em) {};
			}
		}
	}

	\node at ([yshift=1.5 em]l1-3-6-0) {\Large $I_t$};
	\node at ([yshift=1.5 em]l1-3-6-1) {\Large $I_{t+1}$};

	\node[shape=circle, inner sep=0.1 em, fill=black] at (l1-3-3-0) {};

	\node[shape=circle, inner sep=0.1 em, fill=black] at (l1-3-3-1) {};
	\node at ([yshift=-1 em]l1-3-3-0) {$(x, y)$};

	\node at ([yshift=-1 em]l1-3-3-1) {$(x, y)$};

	\node[draw=black, minimum height=3 em, minimum width=3 em, thick] at (l1-3-3-0) {};

	\node at ([yshift=-2 em, xshift=-1.25 em]l1-0-0-1) {$I_{t+1}(x, y) = K(x,y) * P(x,y)$};
    \node at ([yshift=-0.5 em]l1-3-1-0) {$P(x, y)$};

\end{tikzpicture}
		}
		\caption{Kernel-based.}
		\label{subfig:resampling_kernel}
	\end{subfigure}	
	\caption{Representation of transformation-based approaches. (a) Vector-based with a bilinear interpolation. (b) Applying transformations as a convolutional operation. Figure inspired by \cite{Reda2018}.}
	\label{fig:resampling}
\end{figure}

Let ${\mathbf{X}=(X_{t-n},\ldots, X_{t-1}, X_t)}$ be a video sequence of $n$ frames, where $t$ denotes time. Instead of learning the visual appearance, transformation-based approaches assume that visual information is already available in the input sequence. To deal with the strong similarity and pixel redundancy between successive frames, these methods explicitly model the transformations that takes a frame at time $t$ to the frame at $t+1$. These models are formally defined as follows: 
\begin{equation}
\mathbf{Y}_{t+1}=\mathcal{T}\left(\mathcal{G}\left(\mathbf{X}_{t-n: t}\right), \mathbf{X}_{t-n: t}\right),
\end{equation}
where $\mathcal{G}$ is a learned function that outputs future transformation parameters, which applied to the last observed frame $\mathbf{X}_t$ using the function $\mathcal{T}$, generates the future frame prediction $\mathbf{Y}_{t+1}$. According to the classification of Reda \textit{et al.}~\cite{Reda2018}, $\mathcal{T}$ function can be defined as a vector-based resampling such as bilinear sampling, or adaptive kernel-based resampling, e.g. using convolutional operations. For instance, a bilinear sampling operation is defined as: 
\begin{equation}
\mathbf{Y}_{t+1}(x, y)=f\left(\mathbf{X}_{t}(x+u, y+v)\right),
\end{equation} 
where $f$ is a bilinear interpolator such as \cite{Memisevic2010,Memisevic2011,Liu2017}, $(u,v)$ is a motion vector predicted by $\mathcal{G}$, and $X_t(x,y)$ is a pixel value at (x,y) in the last observed frame $X_t$. Approaches following this formulation are categorized as vector-based resampling operations and are depicted in Figure~\ref{subfig:resampling_vector}. 

On the other side, in the kernel-based resampling, the $\mathcal{G}$ function predicts the kernel $\mathrm{K}(x, y)$ which is applied as a convolution operation using $\mathcal{T}$, as depicted in Figure~\ref{subfig:resampling_kernel} and is mathematically represented as follows:
\begin{equation}
\mathbf{Y}_{t+1}(x, y)=\mathrm{K}(x, y) * \mathbf{P}_{t}(x, y),
\end{equation} 
where $\mathrm{K}(x, y)\in\mathbb{R}^{NxN}$ is the 2D kernel predicted by the function $\mathcal{G}$ and $P_t(x,y)$ is an $N \times N$ patch centered at $(x,y)$. 

Combining kernel and vector-based resampling into a hybrid solution, Reda \textit{et al.}~\cite{Reda2018} proposed the \gls{sdc} module that synthesizes high-resolution images applying a learned per-pixel motion vector and kernel at a displaced location in the source image. Their 3D \gls{cnn} model trained on synthetic data and featuring the \gls{sdc} modules, reported promising predictions of a high-fidelity.

\subsubsection{Vector-based Resampling}
Bilinear models use multiplicative interactions to extract transformations from pairs of observations in order to relate images, such as \glspl{gae}~\cite{Memisevic2013a}. Inspired by these models, Michalski \textit{et al.} proposed the \gls{pgp}~\cite{Michalski2014} consisting on a recurrent pyramid of stacked \glspl{gae}. To the best of our knowledge, this was the first attempt to predict future frames in the affine transform space. Multiple \glspl{gae} are stacked to represent a hierarchy of transformations and capture higher-order dependencies. From the experiments on predicting frequency modulated sin-waves, authors stated that standard \glspl{rnn} were outperformed in terms of accuracy. However, no performance comparison was conducted on videos. 
\begin{figure}[tbp]
	\centering
	\includegraphics[width=\linewidth]{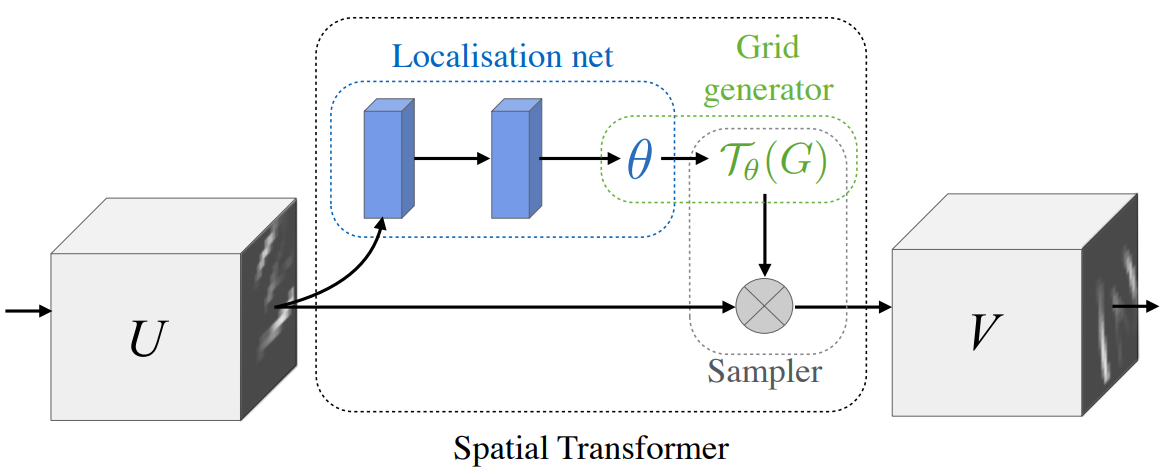}
	\caption{A representation of the spatial transformer module proposed by \cite{Jaderberg2015}. First, the localization network regresses the transformation parameters, denoted as $\theta$, from the input feature map $U$. Then, the grid generator creates a sampling grid from the predicted transformation parameters. Finally, the sampler produces the output map by sampling the input at the points defined in the sampling grid. Figure extracted from~\cite{Jaderberg2015}.}
	\label{fig:spatial_transformer}
\end{figure}

\vspace*{0.3cm}\noindent\textbf{Based on the \gls{st} module \cite{Jaderberg2015}}: To provide spatial transformation capabilities to existing \glspl{cnn}, Jaderberg \textit{et al.}~\cite{Jaderberg2015} proposed the \gls{st} module represented in Figure~\ref{fig:spatial_transformer}. It regresses different affine transformation parameters for each input, to be applied as a single transformation to the whole feature map(s) or image(s). Moreover, it can be incorporated at any part of the \glspl{cnn} and it is fully differentiable. The \gls{st} module is the essence of vector-based resampling approaches for video prediction. As an extension, Patraucean \textit{et al.}~\cite{Patraucean2015} modified the grid generator to consider per-pixel transformations instead of a single dense transformation map for the entire image. They nested a \gls{lstm}-based temporal encoder into a spatial \gls{ae}, proposing the AE-convLSTM-flow architecture. The prediction is generated by resampling the current frame with the flow-based predicted transformation. Using the components of the AE-convLSTM-flow architecture, Lu \textit{et al.}~\cite{Lu2017} assembled an extrapolation module which is unfolded in time for multi-step prediction. Their \gls{fstn} features a novel loss function using the DeePSiM perceptual loss~\cite{Dosovitskiy2016} in order to mitigate blurriness. An exhaustive experimentation and ablation study was carried out, testing multiple combinations of loss functions. Also inspired on the \gls{st} module for the volume sampling layer, Liu \textit{et al.} proposed the \gls{dvf} architecture~\cite{Liu2017}. It consists of a multi-scale flow-based \gls{ed} model originally designed for the video frame interpolation task, but also evaluated on a predictive basis reporting sharp results. Liang \textit{et al.}~\cite{Liang2017} use a flow-warping layer based on a bilinear interpolation. Finn \textit{et al.} proposed the \gls{stp} motion-based model~\cite{Finn2016} producing 2D affine transformations for bilinear sampling. Pursuing efficiency, Amersfoort \textit{et al.}~\cite{Amersfoort2017} proposed a \gls{cnn} designed to predict local affine transformations of overlapping image patches. Unlike the \gls{st} module, authors estimated transformations of input frames off-line and at patch level. As the model is parameter-efficient, it was unfolded in time for multi-step prediction. This resembles \glspl{rnn} as the parameters are shared over time and the local affine transforms play the role of recurrent states.

\subsubsection{Kernel-based Resampling}
As a promising alternative to the vector-based resampling, recent approaches synthesize pixels by convolving input patches with a predicted kernel. However, convolutional operations are limited in learning spatial invariant representations of complex transformations. Moreover, due to their local receptive fields, global spatial information is not fully preserved. Using larger kernels would help to preserve global features, but in exchange for a higher memory consumption. Pooling layers are another alternative, but loosing spatial resolution. Preserving spatial resolution at a low computational cost is still an open challenge for future video frame prediction task. Transformation layers used in vector-based resampling~\cite{Jaderberg2015,Patraucean2015,Liu2017} enabled \glspl{cnn} to be spatially invariant and also inspired kernel-based architectures.

\vspace*{0.3cm}\noindent\textbf{Inspired on the \gls{cdna} module \cite{Finn2016}}: In addition to the \gls{stp} vector-based model, Finn \textit{et al.}~\cite{Finn2016} proposed two different kernel-based motion prediction modules outperforming previous approaches~\cite{Mathieu2016,Oh2015}, (1) the \gls{dna} module predicting different distributions for each pixel and (2) the \gls{cdna} module that instead of predicting different distributions for each pixel, it predicts multiple discrete distributions applied convolutionally to the input. While, \gls{cdna} and \gls{stp} mask out objects that are moving in consistent directions, the \gls{dna} module produces per-pixel motion. These modules inspired several kernel-based approaches. Similar to the \gls{cdna} module, Klein~\textit{et al.} proposed the \gls{dcl}~\cite{Klein2015} for short-range weather prediction. Likewise, Brabandere \textit{et al.}~\cite{Brabandere2016} proposed the \gls{dfn} generating sample (for each image) and position-specific (for each pixel) kernels. This enabled sophisticated and local filtering operations in comparison with the \gls{st} module, that is limited to global spatial transformations. Different to the \gls{cdna} model, the \gls{dfn} uses a softmax layer to filter values of greater magnitude, thus obtaining sharper predictions. Moreover, temporal correlations are exploited using a parameter-efficient recurrent layer, much simpler than~\cite{Srivastava2015,Shi2015}. Exploiting adversarial training, Vondrick \textit{et al.} proposed a \gls{cgan}-based model~\cite{Vondrick2017} consisting of a discriminator similar to~\cite{Vondrick2016} and a \gls{cnn} generator featuring a transformer module inspired on the \gls{cdna} model. Different from the \gls{cdna} model, transformations are not applied recurrently on a per-frame basis. To deal with in-the-wild videos and make predictions invariant to camera motion, authors stabilized the input videos. However, no performance comparison with previous works has been conducted. 

Relying on kernel-based transformations and improving~\cite{Clark2019}, Luc \textit{et al.}~\cite{Luc2020} proposed the \gls{trivdganfp} featuring a novel recurrent unit that computes the parameters of a transformation used to warp previous hidden states without any supervision. These \glspl{tsru} are generic modules and can replace any traditional recurrent unit in currently existent video prediction approaches.

\vspace*{0.3cm}\noindent\textbf{Object-centric representation}: Instead of focusing on the whole input, Chen \textit{et al.}~\cite{Chen2017} modeled individual motion of local objects, i.e. object-centered representations. Based on the \gls{st} module and a pyramid-like sampling~\cite{He2015}, authors implemented an attention mechanism for object selection. Moreover, transformation kernels were generated dynamically as in the \gls{dfn}, to then apply them to the last patch containing an object. Although object-centered predictions is novel, performance drops when dealing with multiple objects and occlusions as the attention module fails to distinguish them correctly.


\subsection{Explicit Motion from Content Separation}
\label{subsc:two_stream}
Drawing inspiration from two-stream architectures for action recognition \cite{Simonyan2014}, video generation from a static image \cite{Vondrick2016}, and unconditioned video generation \cite{Tulyakov2018}, authors decided to factorize the video into content and motion to process each on a separate pathway. By decomposing the high-dimensional videos, the prediction is performed on a lower-dimensional temporal dynamics separately from the spatial layout. Although this makes end-to-end training difficult, factorizing the prediction task into more tractable problems demonstrated good results. 
\begin{figure}[tbp]
	\centering
	\includegraphics[width=0.8\linewidth]{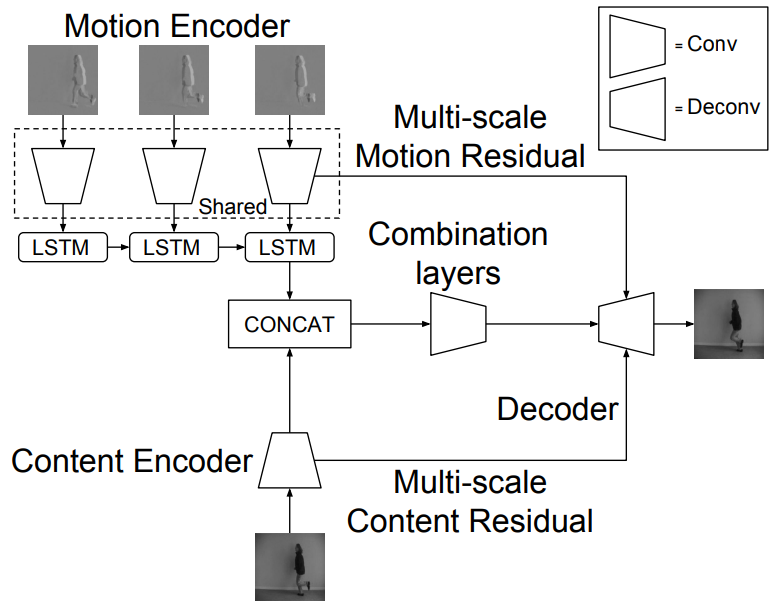}
	\caption{\acrshort{mcnet} with Multi-scale Motion-Content Residuals. While the motion encoder captures the temporal dynamics in a sequence of image differences, the content encoder extracts meaningful spatial features from the last observed RGB frame. After that, the network computes motion-content features that are fed into the decoder to predict the next frame. Figure extracted from \cite{Villegas2017a}.}
	\label{fig:mcnet}
\end{figure}

The \gls{mcnet}~\cite{Villegas2017a}, represented in Figure~\ref{fig:mcnet} was the first end-to-end model that disentangled scene dynamics from the visual appearance. Authors performed an in-depth performance analysis ensuring the motion and content separation through generalization capabilities and stable long-term predictions compared to models that lack of explicit motion-content factorization \cite{Srivastava2015,Mathieu2016}. In a similar fashion, yet working in a higher-level pose space, Denton \textit{et al.}~proposed \gls{drnet}~\cite{Denton2017} using a novel adversarial loss ---it isolates the scene dynamics from the visual content, considered as the discriminative component--- to completely disentangle motion dynamics from content. Outperforming~\cite{Mathieu2016,Villegas2017a}, the \gls{drnet} demonstrated a clean motion from content separation by reporting plausible long-term predictions on both synthetic and natural videos. To improve prediction variability, Liang \textit{et al.}~\cite{Liang2017} fused the future-frame and future-flow prediction into a unified architecture with a shared probabilistic motion encoder. Aiming to mitigate the ghosting effect in disoccluded regions, Gae \textit{et al.}~\cite{Gao2019} proposed a two-staged approach consisting of a separate computation of flow and pixel predictions. As they focused on inpainting occluded regions of the image using flow information, they improved results on disoccluded areas avoiding undesirable artifacts and enhancing sharpness. Separating the moving objects and the static background, Wu \textit{et al.}~\cite{Wu2020} proposed a two-staged architecture that firstly predicts the static background to then, using this information, predict the moving objects in the foreground. Final results are generated through composition and by means of a video inpainting module. Reported predictions are quite accurate, yet performance was not contrasted with the latest video prediction models.

Although previous approaches disentangled motion from content, they have not performed an explicit decomposition into low-dimensional components. Addressing this issue, Hsieh \textit{et al.} proposed the \gls{ddpae}~\cite{Hsieh2018} that decomposes the high-dimensional video into components represented with low-dimensional temporal dynamics. On the Moving MNIST dataset, \gls{ddpae} first decomposes images into individual digits (components) to then factorize each digit into its visual appearance and spatial location, being the latter easier to predict. Although experiments were performed on synthetic data, this approach represents a promising baseline to disentangle and decompose natural videos. Moreover, it is applicable to other existing models to improve their predictions.

\subsection{Conditioned on Extra Variables}
\label{subsec:m_condition}
Conditioning the prediction on extra variables such as vehicle odometry or robot state, among others, would narrow the prediction space. These variables have a direct influence on the dynamics of the scene, providing valuable information that facilitates the prediction task. For instance, the motion captured by a camera placed on the dashboard of an autonomous vehicle is directly influenced by the wheel-steering and acceleration. Without explicitly exploiting this information, we rely blindly on the model's capabilities to correlate the wheel-steering and acceleration with the perceived motion. However, the explicit use of these variables would guide the prediction. 

Following this paradigm, Oh \textit{et al.} first made long-term video predictions conditioned by control inputs from Atari games~\cite{Oh2015}. Although the proposed \gls{ed}-based models reported very long-term predictions (+100), performance drops when dealing with small objects (e.g. bullets in Space Invaders) and while handling stochasticity due to the squared error. However, by simply minimizing $\ell_2$ error can lead to accurate and long-term predictions for deterministic synthetic videos, such as those extracted from Atari video games. Building on~\cite{Oh2015}, Chiappa \textit{et al.}~\cite{Chiappa2017} proposed alternative architectures and training schemes alongside an in-depth performance analysis for both short and long-term prediction. Similar model-based control from visual inputs performed well in restricted scenarios~\cite{Fragkiadaki2016}, but was inadequate for unconstrained environments. These deterministic approaches are unable to deal with natural videos in the absence of control variables.

To address this limitation, the models proposed by Finn et al.~\cite{Finn2016} successfully made predictions on natural images, conditioned on the robot state and robot-object interactions performed in a controlled scenario. These models predict per-pixel transformations conditioned by the previous frame, to finally combine them using a composition mask. They outperformed~\cite{Oh2015,Mathieu2016} on both conditioned and unconditioned predictions, however the quality of long-term predictions degrades over time because of the blurriness caused by the \gls{mse} loss function. Also, using high-dimensional sensory such as images, Dosovitskiy \textit{et al.}~\cite{Dosovitskiy2017} proposed a sensorimotor control model which enables interaction in complex and dynamic \gls{3d} environments. The approach is a \gls{rl}-based techniques, with the difference that instead of building upon a monolithic state and a scalar reward, the authors consider high-dimensional input streams, such as raw visual input, alongside a stream of measurements or player statistics. Although the outputs are future measurements instead of visual predictions, it was proven that using multivariate data benefits decision-making over conventional scalar reward approaches. 

\subsection{In the High-level Feature Space}
\label{subsec:m_high_level}
Despite the vast work on video prediction models, there is still room for improvement in natural video prediction. To deal with the curse of dimensionality, authors reduced the prediction space to high-level representations, such as semantic and instance segmentation, and human pose. Since the pixels are categorical, the semantic space greatly simplifies the prediction task, yet unexpected deformations in semantic maps and disocclusions, i.e. initially occluded scene entities become visible, induce uncertainty.
However, high-level prediction spaces are more tractable and constitute good intermediate representations. By bypassing the prediction in the pixel space, models become able to report longer-term and more accurate predictions.

\subsubsection{Semantic Segmentation}
In recent years, semantic and instance representations have gained increasing attention, emerging as a promising avenue for complete scene understanding. By decomposing the visual scene into semantic entities, such as pedestrians, vehicles and obstacles, the output space is narrowed to high-level scene properties. This intermediate representation represents a more tractable space as pixel values of a semantic map are categorical. In other words, scene dynamics are modeled at the semantic entity level instead of being modeled at the pixel level. This has encouraged authors to (1) leverage future prediction to improve parsing results~\cite{Jin2017} and (2) directly predict segmentation maps into the future~\cite{Luc2017, Luc2018, Luc2019}.
\begin{figure*}[tbp]
	\centering
	\includegraphics[width=0.9\linewidth]{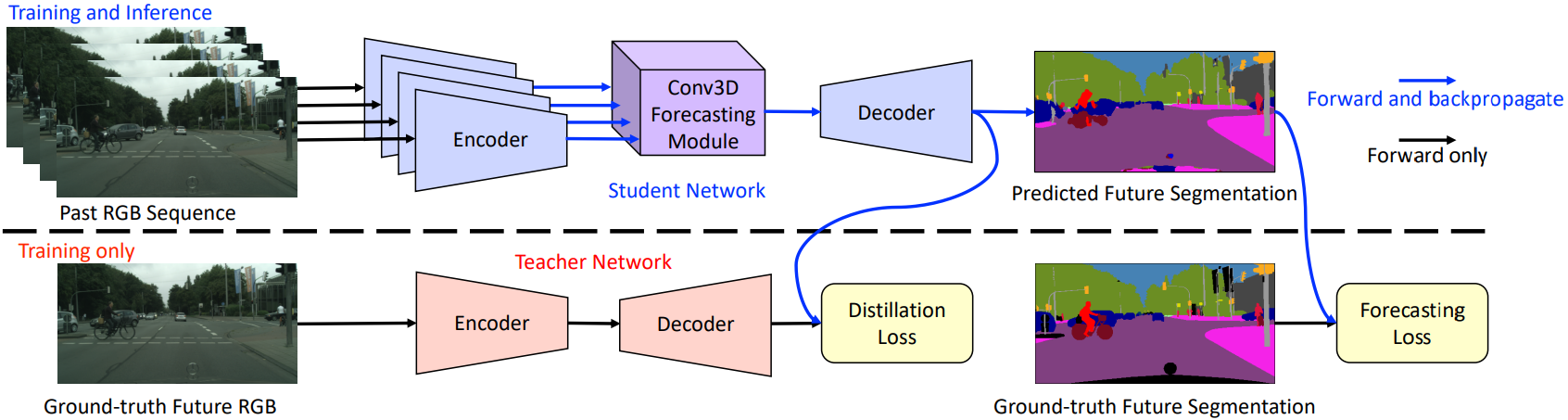}
	\caption{Two-staged method proposed by Chiu \textit{et al.} \cite{Chiu2019}. In the upper half, the student network consists on an \gls{ed}-based architecture featuring a 3D convolutional forecasting module. It performs the forecasting task guided by an additional loss generated by the teacher network (represented in the lower half). Figure extracted from \cite{Chiu2019}.}
	\label{fig:student_teacher}
\end{figure*}

Exploring the scene parsing in future frames, Jin \textit{et al.} proposed the \gls{pearl} framework~\cite{Jin2017} which was the first to explore the potential of a \gls{gan}-based frame prediction model to improve per-pixel segmentation. Specifically, this framework conducts two complementary predictive learning tasks. Firstly, it captures the temporal context from input data by using a single-frame prediction network. Then, these temporal features are embedded into a frame parsing network through a transform layer for generating per-pixel future segmentations. Although the predictive net was not compared with existing approaches, \gls{pearl} outperforms the traditional parsing methods by generating temporally consistent segmentations. In a similar fashion, Luc \textit{et al.}~\cite{Luc2017} extended the ms\gls{cnn} model of~\cite{Mathieu2016} to the novel task of predicting semantic segmentations of future frames, using softmax pre-activations instead of raw pixels as input. The use of intermediate features or higher-level data as input is a common practice in the video prediction performed in the high-level feature space. Some authors refer to this type or input data as percepts. Luc \textit{et al.} explored different combinations of loss functions, inputs (using RGB information alongside percepts), and outputs (autoregressive and batch models). Results on short, medium and long-term predictions are sound, however, the models are not end-to-end and they do not capture explicitly the temporal continuity across frames. To address this limitation and extending~\cite{Jin2017}, Jin \textit{et al.} first proposed a model for jointly predicting motion flow and scene parsing~\cite{Jin2017a}. Flow-based representations implicitly draw temporal correlations from the input data, thus producing temporally coherent per-pixel segmentations. As in~\cite{Luc2017}, the authors tested different network configurations, as using Res101-FCN percepts for the prediction of semantic maps, and also performed multi-step prediction up to 10 time-steps into the future. Per-pixel accuracy improved when segmenting small objects, e.g. pedestrians and traffic signs, which are more likely to vanish in long-term predictions. Similarly, except that time dimension is modeled with \glspl{lstm} instead of motion flow estimation, Nabavi \textit{et al.} proposed a simple bidirectional \gls{ed}-\gls{lstm}~\cite{Nabavi2018} using segmentation masks as input. Although the literature on knowledge distillation~\cite{Ba2014,Hinton2015} stated that softmax pre-activations carry more information than class labels, this model outperforms~\cite{Jin2017a,Luc2017} on short-term predictions.

Another relevant idea is to use both motion flow estimation alongside \gls{lstm}-based temporal modeling. In this direction, Terwilliger \textit{et al.}~\cite{Terwilliger2019} proposed a novel method performing a \gls{lstm}-based feature-flow aggregation. Authors also tried to further simplify the semantic space by disentangling motion from semantic entities~\cite{Villegas2017a}, achieving low overhead and efficiency. The prediction problem was decomposed into two subtasks, that is, current frame segmentation and future optical flow prediction, which are finally combined with a novel end-to-end warp layer. An improvement on short-term predictions were reported over previous works~\cite{Luc2017,Jin2017a}, yet performing worse on mid-term predictions. 

A different approach was proposed by Vora \textit{et al.}~\cite{Vora2018} which first incorporated structure information to predict future 3D segmented point clouds. Their geometry-based model consists of several derivable sub-modules: (1) the pixel-wise segmentation and depth estimation modules which are jointly used to generate the \gls{3d} segmented point cloud of the current RGB frame; and (2) an \gls{lstm}-based module trained to predict future camera ego-motion trajectories. The future \gls{3d} segmented point clouds are obtained by transforming the previous point clouds with the predicted ego-motion. Their short-term predictions improved the results of \cite{Luc2017}, however, the use of structure information for longer-term predictions is not clear.

The main disadvantage of two-staged, i.e. not end-to-end, approaches~\cite{Luc2017,Jin2017a,Nabavi2018,Vora2018,Terwilliger2019} is that their performance is constrained by external supervisory signals, e.g. optical flow~\cite{Revaud2015}, segmentation~\cite{Zhao2017a} and intermediate features or percepts~\cite{Yu2017}. Breaking this trend, Chiu \textit{et al.}~\cite{Chiu2019} first solved jointly the semantic segmentation and forecasting problems in a single end-to-end trainable model by using raw pixels as input. This \gls{ed} architecture is based on two networks, with one performing the forecasting task (student) and the other (teacher) guiding the student by means of a novel knowledge distillation loss. An in-depth ablation study was performed, validating the performance of the \gls{ed} architectures as well as the 3D convolution used for capturing temporal scale instead of a \gls{lstm} or \gls{convlstm}, as in previous works. 

Avoiding the flood of deterministic models, Bhattacharyya \textit{et al.} proposed a Bayesian formulation of the ResNet model in a novel architecture to capture model and observation uncertainty \cite{Bhattacharyya2019}. As main contribution, their dropout-based Bayesian approach leverages synthetic likelihoods~\cite{Rosca2017} to encourage prediction diversity and deal with multi-modal outcomes. Since Cityscapes sequences have been recorded in the frame of reference of a moving vehicle, authors conditioned the predictions on vehicle odometry. 

\subsubsection{Instance Segmentation}
While great strides have been made in predicting future segmentation maps, the authors attempted to make predictions at a semantically richer level, i.e. future prediction of semantic instances. Predicting future instance-level segmentations is a challenging and weakly unexplored task. This is because instance labels are inconsistent and variable in number across the frames in a video sequence. Since the representation of semantic segmentation prediction models is of fixed-size, they cannot directly address semantics at the instance level.

To overcome this limitation and introducing the novel task of predicting instance segmentations, Luc \textit{et al.}~\cite{Luc2018} predict fixed-sized feature pyramids, i.e. features at multiple scales, used by the Mask R-CNN~\cite{He2017} network. The combination of dilated convolutions and multi-scale, efficiently preserve high-resolution details improving the results over previous methods~\cite{Luc2017}. To further improve predictions, Sun \textit{et al.}~\cite{Sun2019} focused on modeling not only the spatio-temporal correlations between the pyramids, but also the intrinsic relations among the feature layers inside them. By enriching the contextual information using the proposed Context Pyramid \glspl{convlstm} (CP-ConvLSTM), an improvement in the prediction was noticed. Although the authors have not shown any long-term predictions nor compared with semantic segmentation models, their approach is currently the state of the art in the task of predicting instance segmentations, outperforming~\cite{Luc2018}.

\subsubsection{Other High-level Spaces}
Although semantic and instance segmentation spaces were the most used in video prediction, other high-level spaces such as human pose and keypoints represent a promising avenue.

\vspace*{0.2cm}\noindent\textbf{Human Pose}: As the human pose is a low-dimensional and interpretable structure, it represents a cheap supervisory signal for predictive models. This fostered pose-guided prediction methods, where future frame regression in the pixel space is conditioned by intermediate prediction of human poses. However, these methods are limited to videos with human presence. As this review focuses on video prediction, we will briefly review some of the most relevant methods predicting human poses as an intermediate representation. 

From a supervised prediction of human poses, Villegas \textit{et al.}~\cite{Villegas2017} regress future frames through analogy making~\cite{Reed2015}. Although background is not considered in the prediction, authors compared the model against~\cite{Shi2015,Mathieu2016} reporting long-term results. To make the model unsupervised on the human pose, Wichers \textit{et al.}~\cite{Wichers2018} adopted different training strategies: end-to-end prediction minimizing the $\ell_2$ loss, and through analogy making, constraining the predicted features to be close to the outputs of the future encoder. Different from~\cite{Villegas2017}, in this work the predictions are made in the feature space. As a probabilistic alternative, Walker \textit{et al.}~\cite{Walker2017} fused a \gls{cvae}-based probabilistic pose predictor with a \gls{gan}. While the probabilistic predictor enhances the diversity in the predicted poses, the adversarial network ensures prediction realism. As this model struggles with long-term predictions, Fushishita \textit{et al.}~\cite{Fushishita2019} addressed long-term video prediction of multiple outcomes avoiding the error accumulation and vanishing gradients by using a unidimensional \gls{cnn} trained in an adversarial fashion. To enable multiple predictions, they have used additional inputs ensuring trajectory and behavior variability at a human pose level. To better preserve the visual appearance in the predictions than~\cite{Lee2018,Villegas2017a,Villegas2017}, Tang \textit{et al.}~\cite{Tang2019} firstly predict human poses using a \gls{lstm}-based model to then synthesize pose-conditioned future frames using a combination of different networks: a global \gls{gan} modeling the time-invariant background and a coarse human pose, a local \gls{gan} refining the coarse-predicted human pose, and a 3D-\gls{ae} to ensure temporal consistency across frames. 

\vspace*{0.3cm}\noindent\textbf{Keypoints-based representations}: The keypoint coordinate space is a meaningful, tractable and structured representation for prediction, ensuring stable learning. It enforces model's internal representation to contain object-level information. This leads to better results on tasks requiring object-level understanding such as, trajectory prediction, action recognition and reward prediction. As keypoints are a natural representation of dynamic objects, Minderer \textit{et al.}~\cite{Minderer2019} reformulated the prediction task in the keypoint coordinate space. They proposed an \gls{ae} architecture with a keypoint-based representational bottleneck, consisting of a \gls{vrnn} that predicts dynamics in the keypoint space. Although this model qualitatively outperforms the \gls{svg}~\cite{Denton2018}, \gls{savp}~\cite{Lee2018} and \acrshort{epva}~\cite{Wichers2018} models, the quantitative evaluation reported similar results. 

\begin{table*} \centering
	\ra{1.18}
	\caption{Summary of video prediction models (\textbf{c}: \textbf{c}onvolutional; \textbf{r}: \textbf{r}ecurrent; \textbf{v}: \textbf{v}ariational; \textbf{ms}: \textbf{m}ulti-\textbf{s}cale; \textbf{st}: \textbf{st}acked; \textbf{bi}: \textbf{bi}directional; \textbf{P}: \textbf{P}ercepts; \textbf{M}: \textbf{M}otion; \textbf{PL}: \textbf{P}erceptual \textbf{L}oss; \textbf{AL}: \textbf{A}dversarial \textbf{L}oss; \textbf{S}/\textbf{R}: using \textbf{S}ynthetic/\textbf{R}eal datasets; \textbf{SS}: \textbf{S}emantic \textbf{S}egmentation; \textbf{D}: \textbf{D}epth; \textbf{S}: \textbf{S}tate; \textbf{Po}: \textbf{Po}se; \textbf{O}: \textbf{O}dometry; \textbf{IS}: \textbf{I}nstance \textbf{S}egmentation; \textbf{ms}: \textbf{m}ulti-\textbf{s}tep prediction; \textbf{pred-fr}: number of \textbf{pred}icted \textbf{fr}ames, $\0$ 1-5 frames, $\0\0$ 5-10 frames, $\0\0\0$ 10-100 frames, $\0\0\0\0$ over 100 frames; \textbf{ood}: indicates if model was tested on \textbf{o}ut-\textbf{o}f-\textbf{d}omain tasks).}
	\label{table:methods}
	\resizebox{\textwidth}{!}{
	\begin{tabular} {@{}lccccccccclcc@{}} 
		\toprule
		& & & & & \multicolumn{4}{c}{\textbf{details}} & \multicolumn{3}{c}{\textbf{evaluation}} & \\
		\cmidrule(lr){6-9} \cmidrule(lr){10-12} 
		\textbf{method} & \textbf{year} & \textbf{based on} & \textbf{architecture} & \textbf{datasets (train, valid, test)} & \textbf{input} & \textbf{output} & \textbf{MS} & \textbf{loss function} & \textbf{S/R} & \textbf{pred-fr} & \textbf{ood} & \textbf{code} \\
		\midrule
		\multicolumn{13}{c}{\textbf{Direct Pixel Synthesis}} \\
		\midrule
		Ranzato \textit{et al.} \cite{Ranzato2014} & \num{2014} & \cite{Bengio2003,Mikolov2010} & r\gls{cnn} & \cite{Soomro2012,Cadieu2012} & RGB & RGB & \xmark & $\gls{ce}$ & R & $\0$ & \xmark & \xmark \\ 
		Srivastava \textit{et al.} \cite{Srivastava2015} & \num{2015} & \cite{Sutskever2014} & \gls{lstm}-\gls{ae} & \cite{Soomro2012,Kuehne2011,Srivastava2015,Karpathy2014} & RGB,P & RGB & \checkmark & $\gls{ce},\ell_2$ & SR & $\0\0\0$ & \checkmark & \checkmark \\ 
		\gls{pgn} \cite{Lotter2015} & \num{2015} & - & \gls{lstm}-c\gls{ed} & \cite{Sutskever2008} & RGB & RGB & \xmark & $\gls{mse},AL$ & S & $\0$ &  \xmark & \xmark \\
		Shi \textit{et al.} \cite{Shi2015} & \num{2015} & \cite{Srivastava2015} & c\gls{lstm} & \cite{Srivastava2015} & RGB & RGB & \xmark & $\gls{ce}$ & S & $\0\0\0$ &  \checkmark & \xmark \\ 
		BeyondMSE \cite{Mathieu2016} & \num{2016} & \cite{Denton2015,Mahendran2015} & ms\gls{cnn} & \cite{Soomro2012,Karpathy2014} & RGB & RGB & \checkmark & $\ell_1,\gls{gdl},AL$ & R & $\0\0$ & \xmark & \checkmark \\ 
		PredNet \cite{Lotter2017} & \num{2017} & \cite{Rao1999,Chalasani2013} & st\glspl{lstm} & \cite{Geiger2013,Dollar2009,Ionescu2014,Santana2016} & RGB & RGB & \checkmark & $\ell_1$,$\ell_2$ & SR & $\0\0$ & \checkmark & \checkmark \\ 
		ContextVP \cite{Byeon2018} & \num{2018} & \cite{Graves2007,Stollenga2015} & \gls{mdlstm} &  \cite{Soomro2012,Geiger2013,Dollar2009,Ionescu2014} & RGB & RGB & \checkmark & $\ell_1,\gls{gdl}$ & R & $\0\0$ & \xmark & \xmark \\
		\gls{frnn} \cite{Oliu2018} & \num{2018} & - & cGRU-\gls{ae} & \cite{Srivastava2015,Schuldt2004,Soomro2012} & RGB & RGB & \checkmark & $\ell_1$ & SR & $\0\0\0$ & \xmark & \checkmark \\
		\gls{e3dlstm} \cite{Wang2019b} & \num{2019} & \cite{Shi2015} & r3D-CNN & \cite{Srivastava2015,Schuldt2004,Zhang2017a,Goyal2017} & RGB & RGB & \checkmark & $\ell_1,\ell_2,\gls{ce}$ & SR & $\0\0\0$ & \checkmark & \checkmark \\
		Kwon \textit{et al.} \cite{Kwon2019} & \num{2019} & \cite{Yi2017,Zhu2017,Johnson2016} & cycleGAN & \cite{Geiger2013,Dollar2009,Soomro2012,Luo2017a,Ravanbakhsh2017} & RGB & RGB & \checkmark & $\ell_1,\acrshort{log},AL$ & R & $\0\0\0$ & \xmark & \xmark \\
		Znet \cite{Zhang2019} & \num{2019} & \cite{Shi2015} & c\gls{lstm} & \cite{Srivastava2015,Schuldt2004} & RGB & RGB & \checkmark & $\ell_2,BCE,AL$ & SR & $\0\0\0$ & \xmark & \xmark \\
		VPGAN \cite{Hu2019} & \num{2019} & \cite{Zhu2017,Denton2017} & \gls{gan} & \cite{Ebert2017,Schuldt2004} & RGB,Z & RGB & \checkmark & $\ell_1,L_{cycle},AL$ & R & $\0\0\0$ & \xmark & \xmark \\
		Jin \textit{et al.}\cite{Jin2020} & \num{2020} & - & c\gls{ed}-\gls{gan} & \cite{Schuldt2004,Dollar2009,Geiger2013,Ebert2017} & RGB & RGB & \checkmark & $\ell_2,\gls{gdl},AL$ & R & $\0\0\0$ & \xmark & \xmark \\
		Shouno \textit{et al.}\cite{Shouno2020} & \num{2020} & \cite{Lotter2017} & \gls{gan} & \cite{Dollar2009,Geiger2013} & RGB & RGB & \checkmark & $L_p,AL,PL$ & R & $\0\0\0$ & \xmark & \xmark \\
		\gls{crevnet}\cite{Yu2020} & \num{2020} & \cite{Dinh2015,Wang2017,Shi2015} & 3d-c\gls{ed} & \cite{Srivastava2015,Dollar2009,Geiger2013,traffic4cast} & RGB & RGB & \checkmark & $\gls{mse}$ & SR & $\0\0\0$ & \checkmark & \checkmark \\
		\midrule
		\multicolumn{13}{c}{\textbf{Using Explicit Transformations}} \\
		\midrule
		\gls{pgp} \cite{Michalski2014} & \num{2014} & \cite{Memisevic2011} & st-r\glspl{gae} & \cite{Memisevic2013,Sutskever2008} & RGB & RGB & \checkmark & $\ell_2$ & SR & $\0$ & \xmark & \xmark \\ 
		Patraucean \textit{et al.} \cite{Patraucean2015} & \num{2015} & \cite{Sutskever2014} & \gls{lstm}-c\gls{ae} & \cite{Srivastava2015,Kuehne2011,Santner2010,Vezzani2010} & RGB & RGB & \xmark & $\ell_2,\ell_\delta$ & SR & $\0$ & \checkmark & \checkmark \\
		\gls{dfn} \cite{Brabandere2016} & \num{2016} & \cite{Klein2015,Finn2016} & r-c\gls{ed} & \cite{Soomro2012,Srivastava2015} & RGB & RGB & \checkmark & $B\gls{ce}$ & SR & $\0\0\0$ & \checkmark & \checkmark \\
		Amersfoort \textit{et al.} \cite{Amersfoort2017} & \num{2017} & \cite{Patraucean2015} & \gls{cnn} & \cite{Soomro2012,Srivastava2015} & RGB & RGB & \checkmark & $\gls{mse}$ & SR & $\0\0$ & \xmark & \xmark \\
		\gls{fstn} \cite{Lu2017} & \num{2017} & \cite{Patraucean2015,Dosovitskiy2016} & \gls{lstm}-c\gls{ed} & \cite{Soomro2012,Srivastava2015,Santner2010,Vezzani2010,Karpathy2014} & RGB & RGB & \checkmark & $\ell_2,\ell_\delta,PL$ & SR & $\0\0\0$ & \xmark & \xmark \\
		Vondrick \textit{et al.} \cite{Vondrick2017} & \num{2017} & \cite{Finn2016,Vondrick2016} & \gls{cgan} & \cite{Thomee2016} & RGB & RGB & \checkmark & $\gls{ce},AL$ & R & $\0\0\0$ & \checkmark & \xmark \\
		Chen \textit{et al.} \cite{Chen2017} & \num{2017} & \cite{Amersfoort2017,Jaderberg2015,Brabandere2016} & rCNN-ED & \cite{Soomro2012,Srivastava2015} & RGB & RGB & \checkmark & $CE,\ell_2,\gls{gdl},AL$ & SR & $\0\0$ & \xmark & \xmark \\
		\gls{dvf} \cite{Liu2017} & \num{2017} & \cite{Jaderberg2015} & ms-cED & \cite{Soomro2012,Idrees2017} & RGB & RGB & \checkmark & $\ell_1, \acrshort{tv}$ & R & $\0$ & \checkmark & \checkmark \\
		SDC-Net \cite{Reda2018} & \num{2018} & \cite{Niklaus2017,Niklaus2017a} & \gls{cnn} & \cite{Dollar2009,Abu-El-Haija2016} & RGB,M & RGB & \checkmark & $\ell_1,PL$ & SR & $\0\0$ & \checkmark & \xmark \\
		TrIVD-GAN-FP \cite{Luc2020} & \num{2020} & \cite{Clark2019,Hao2018,Gao2019} & DVD-GAN & \cite{Soomro2012,Ebert2017,Carreira2018} & RGB & RGB & \checkmark & $L_{hinge}$\cite{Liang2017} & R & $\0\0\0$ & \xmark & \xmark \\
		\midrule
		\multicolumn{13}{c}{\textbf{Explicit Motion from Content Separation}} \\
		\midrule
		\gls{mcnet} \cite{Villegas2017a} & \num{2017} & \cite{Shi2015,Xue2016,Simonyan2014} & \gls{lstm}-c\gls{ed} & \cite{Soomro2012,Karpathy2014,Gorelick2007,Schuldt2004} & RGB & RGB & \checkmark & $\ell_p,\gls{gdl},AL$ & R & $\0\0\0$ & \xmark & \checkmark \\
		Dual-GAN \cite{Liang2017} & \num{2017} & \cite{Goodfellow2014} & \gls{vae}-\gls{gan} & \cite{Soomro2012,Dollar2009,Geiger2013,Idrees2017} & RGB & RGB & \checkmark & $\ell_1,KL,AL$ & R & $\0\0$ & \xmark & \xmark \\
		\gls{drnet} \cite{Denton2017} & \num{2017} & \cite{Villegas2017a} & \gls{lstm}-\gls{ed} & \cite{Srivastava2015,LeCun2004,Song2017,Schuldt2004} & RGB & RGB & \checkmark & $\ell_2,\gls{ce},AL$ & SR & $\0\0\0\0$ & \checkmark & \checkmark \\
		DPG \cite{Gao2019} & \num{2019} & \cite{Finn2016,Hao2018} & c\gls{ed} & \cite{Dollar2009,Menze2015,Janai2018} & RGB & RGB & \checkmark & $\ell_p,\gls{tv},PL,CE$ & SR & $\0\0$ & \xmark & \xmark \\
		\midrule
		\multicolumn{13}{c}{\textbf{Conditioned on Extra Variables}} \\
		\midrule
		Oh \textit{et al.} \cite{Oh2015} & \num{2015} & \cite{Shi2015} & r\gls{ed} & \cite{Bellemare2013} & RGB,A & RGB & \checkmark & $\ell_2$ & S & $\0\0\0\0$  & \checkmark & \checkmark \\ 
		Finn \textit{et al.} \cite{Finn2016} & \num{2016} & \cite{Shi2015,Oh2015} & st-c\glspl{lstm} & \cite{Finn2016,Ionescu2014} & RGB,A,S & RGB & \checkmark & $\ell_2$ & R & $\0\0\0$ & \xmark & \checkmark \\ 
		\midrule
		\multicolumn{13}{c}{\textbf{In the High-level Feature Space}} \\
		\midrule
		Villegas \textit{et al.} \cite{Villegas2017} & \num{2017} & \cite{Reed2015,Reed2016,Newell2016} & \gls{lstm}-c\gls{ed} & \cite{Ionescu2014,Zhang2013} & RGB,Po & RGB,Po & \checkmark & $\ell_2,PL,AL$\cite{Dosovitskiy2016} & R & $\0\0\0\0$ & \checkmark & \xmark \\
		\gls{pearl} \cite{Jin2017} & \num{2017} & - & c\gls{ed} & \cite{Cordts2016,Brostow2008} & RGB & SS & \xmark & $\ell_2,AL$ & R & $\0$ & \checkmark & \xmark \\ 
		S2S \cite{Luc2017} & \num{2017} & \cite{Mathieu2016} & ms\gls{cnn} & \cite{Cordts2016,Brostow2008} & P & SS & \checkmark & $\ell_1,\gls{gdl},AL$ & R & $\0\0\0$ & \xmark & \checkmark \\ 
		Walker \textit{et al.} \cite{Walker2017} & \num{2017} & \cite{Fragkiadaki2015} & v\gls{ed} & \cite{Soomro2012,Zhang2013} & RGB,Po & RGB & \checkmark & $\ell_2,\gls{ce},KL,AL$ & R & $\0\0\0$ & \checkmark & \xmark  \\ 
		Jin et al. \cite{Jin2017a} & \num{2017} & \cite{Jin2017,Luc2017,Patraucean2015} & c\gls{ed} & \cite{Cordts2016,Santana2016} & RGB,P & SS,M & \checkmark & $\ell_1,\gls{gdl},\gls{ce}$ & R & $\0\0\0$ & \checkmark & \xmark \\ 
		\gls{epva} \cite{Wichers2018} & \num{2018} & \cite{Villegas2017} & \gls{lstm}-\gls{ed} & \cite{Ionescu2014} & RGB & RGB & \checkmark & $\ell_2,AL$ & SR & $\0\0\0\0$ & \checkmark & \checkmark \\ 
		Nabavi \textit{et al.} \cite{Nabavi2018} & \num{2018} & \cite{Jin2017a,Luc2017} & bi\gls{lstm}-c\gls{ed} & \cite{Cordts2016} & P & SS & \checkmark & $\gls{ce}$ & R & $\0\0$ & \xmark & \xmark \\
		F2F \textit{et al. }\cite{Luc2018} & \num{2018} & \cite{He2017,Luc2017} & st-ms\gls{cnn} & \cite{Cordts2016} & P & P,SS,IS & \checkmark & $\ell_2$ & R & $\0\0\0$ & \checkmark & \checkmark \\
		Vora \textit{et al.} \cite{Vora2018} & \num{2018} & - & \gls{lstm} & \cite{Cordts2016} & ego-M & ego-M & \xmark & $\ell_1$ & R & $\0$ & \checkmark & \xmark \\ 
		Chiu \textit{et al.} \cite{Chiu2019} & \num{2019} & - & 3D-c\gls{ed} & \cite{Cordts2016,Huang2018a} & RGB & SS & \xmark & $\gls{ce},\gls{mse}$ & R & $\0\0$ & \xmark & \xmark \\
		Bayes-WD-SL \cite{Bhattacharyya2019} & \num{2019} & \cite{Luc2017,Jin2017a} & bayesResNet & \cite{Cordts2016} & SS,O & SS & \checkmark & $KL$ & SR & $\0\0\0$ & \checkmark & \checkmark \\
		Sun \textit{et al.} \cite{Sun2019} & \num{2019} & \cite{Luc2018} & st-ms-c\gls{lstm} & \cite{Cordts2016,Seguin2016} & P & P,IS & \checkmark & $\ell_2$,\cite{He2017} & R & $\0\0$ & \xmark & \xmark  \\ 
		Terwilliger \textit{et al.} \cite{Terwilliger2019} & \num{2019} & \cite{Jin2017a,Villegas2017a} & M-c\gls{lstm} & \cite{Cordts2016} & RGB,P & SS & \checkmark & $\gls{ce},\ell_1$ & R & $\0\0\0$ & \xmark & \checkmark \\
		Struct-\gls{vrnn} \cite{Minderer2019} & \num{2019} & \cite{Jakab2018,Zhang2018} & c\gls{vrnn} & \cite{Zhan2019,Ionescu2014} & RGB & RGB & \checkmark & $\ell_2,KL$ & SR & $\0\0$ & \checkmark & \checkmark \\
		\midrule
		\multicolumn{13}{c}{\textbf{Incorporating Uncertainty}} \\
		\midrule
		Goroshin \textit{et al.} \cite{Goroshin2015a} & \num{2015} & \cite{Hinton2011} & c\gls{ae} & \cite{LeCun2004,Goroshin2015} & RGB & RGB & \xmark & $\ell_2, penalty$ & SR & $\0$ & \xmark & \xmark \\
		Fragkiadaki \textit{et al.} \cite{Fragkiadaki2017} & \num{2017} & \cite{Walker2016,Xue2016} & v\gls{ed} & \cite{Ionescu2014,Brox2010} & RGB & RGB & \xmark & $KL,MCbest$ & R & $\0$ & \checkmark & \xmark \\
		\acrshort{een} \cite{Henaff2017} & \num{2017} & \cite{Rao1999,Lotter2017,Schmidhuber1992} & v\gls{ed} & \cite{Agrawal2016,Mnih2016,Zhang2017} & RGB & RGB & \checkmark & $\ell_1,\ell_2$ & SR & $\0\0$ & \xmark & \checkmark \\ 
		\acrshort{sv2p} \cite{Babaeizadeh2018} & \num{2018} & \cite{Finn2016} & \gls{cdna} & \cite{Ebert2017,Ionescu2014,Finn2016} & RGB & RGB & \checkmark & $\ell_p,KL$ & SR & $\0\0\0$ & \xmark & \checkmark \\
		\gls{svg} \cite{Denton2018} & \num{2018} & \cite{Babaeizadeh2018} & \gls{lstm}-c\gls{ed} & \cite{Srivastava2015,Schuldt2004,Ebert2017} & RGB & RGB & \checkmark & $\ell_2,KL$ & SR & $\0\0\0\0$ & \xmark & \checkmark \\ 
		Castrejon \textit{et al.} \cite{Castrejon2019} & \num{2019} & \cite{Denton2018,Chung2015} & v\gls{rnn} & \cite{Ebert2017,Srivastava2015,Cordts2016} & RGB & RGB & \checkmark & $KL$ & SR & $\0\0\0$ & \xmark & \xmark \\
		Hu \textit{et al.}\cite{Hu2020} & \num{2020} & \cite{Clark2019,Luc2017,Kohl2018} & c\gls{ed} & \cite{Cordts2016,Huang2018a,Yu2018,Neuhold2017} & RGB & SS,D,M & \checkmark & $\gls{ce},\ell_\delta,L_d,L_c,L_p$ & R & $\0\0\0$ & \checkmark & \xmark \\
		\bottomrule
	\end{tabular}}
\end{table*}

\subsection{Incorporating Uncertainty}
\label{subsec:m_uncertainty}
Although high-level representations significantly reduce the prediction space, the underlying distribution still has multiple modes. In other words, different plausible outcomes would be equally probable for the same input sequence. Addressing multimodal distributions is not straightforward for regression and classification approaches, as they regress to the mean and aim to discretize a continuous high-dimensional space, respectively. To deal with the inherent unpredictability of natural videos, some works introduced latent variables into existing deterministic models or directly relied on generative models such as \glspl{gan} and \glspl{vae}.

Inspired by the \gls{dvf}, Xue \textit{et al.}~\cite{Xue2016} proposed a \gls{cvae}-based~\cite{Kingma2014,Yan2016} multi-scale model featuring a novel cross convolutional layer trained to regress the difference image or Eulerian motion~\cite{Wu2012}. Background on natural videos is not uniform, however the model implicitly assumes that the difference image would accurately capture the movement in foreground objects. Introducing latent variables into a convolutional \gls{ae}, Goroshin \textit{et al.}~\cite{Goroshin2015a} proposed a probabilistic model for learning linearized feature representations to linearly extrapolate the predicted frame in a feature space. Uncertainty is introduced to the loss by using a cosine distance as an explicit curvature penalty. Authors focused on evaluating the linearization properties, yet the model was not contrasted to previous works. Extending~\cite{Xue2016,Walker2016}, Fragkiadaki \textit{et al.}~\cite{Fragkiadaki2017} proposed several architectural changes and training schemes to handle marginalization over stochastic variables, such as sampling from the prior and variational inference. They proposed a stochastic \gls{ed} architecture that predicts future optical flow, i.e., dense pixel motion field, used to spatially transform the current frame into the next frame prediction. To introduce uncertainty in predictions, the authors proposed the k-best-sample-loss (MCbest) that draws $K$ outcomes penalizing those similar to the ground-truth. 

Incorporating latent variables into the deterministic \gls{cdna} architecture for the first time, Babaeizadeh \textit{et al.} proposed the \gls{sv2p}~\cite{Babaeizadeh2018} model handling natural videos. Their time-invariant posterior distribution is approximated from the entire input video sequence. Authors demonstrated that, by explicitly modeling uncertainty with latent variables, the deterministic \gls{cdna} model is outperformed. By combining a standard deterministic architecture (\gls{lstm}-\gls{ed}) with stochastic latent variables, Denton \textit{et al.} proposed the \gls{svg} network~\cite{Denton2018}. Different from \gls{sv2p}, the prior is sampled from a time-varying posterior distribution, i.e. it is a learned-prior instead of fixed-prior sampled from the same distribution. Most of the \glspl{vae} use a fixed Gaussian as a prior, sampling randomly at each time step. Exploiting the temporal dependencies, a learned-prior predicts high variance in uncertain situations, and a low variance when a deterministic prediction suffices. The \gls{svg} model is easier to train and reported sharper predictions in contrast to~\cite{Babaeizadeh2018}. Built upon \gls{svg}, Villegas \textit{et al.} \cite{Villegas2019} implemented a baseline to perform an in-depth empirical study on the importance of the inductive bias, stochasticity, and model's capacity in the video prediction task. Different from previous approaches, Henaff \textit{et al.} proposed the \gls{een}~\cite{Henaff2017} that incorporates uncertainty by feeding back the residual error ---the difference between the ground truth and the deterministic prediction--- encoded as a low-dimensional latent variable. In this way, the model implicitly separates the input video into deterministic and stochastic components.

On the one hand, latent variable-based approaches cover the space of possible outcomes, yet predictions lack of realism. On the other hand, \glspl{gan} struggle with uncertainty, but predictions are more realistic. Searching for a trade-off between \glspl{vae} and \glspl{gan}, Lee \textit{et al.}~\cite{Lee2018} proposed the \gls{savp} model, being the first to combine latent variable models with \glspl{gan} to improve variability in video predictions, while maintaining realism. Under the assumption that blurry predictions of \glspl{vae} are a sign of underfitting, Castrejon \textit{et al.} extended the VRNNs to leverage a hierarchy of latent variables and better approximate data likelihood~\cite{Castrejon2019}. Although the backpropagation through a hierarchy of conditioned latents is not straightforward, several techniques alleviated this issue such as, KL beta warm-up, dense connectivity pattern between inputs and latents, \glspl{lvae} \cite{Sonderby2016}. As most of the probabilistic approaches fail in approximating the true distribution of future frames, Pottorff \textit{et al.}~\cite{Pottorff2019} reformulated the video prediction task without making any assumption about the data distribution. They proposed the \gls{ile} enabling exact maximum likelihood learning of video sequences, by combining an invertible neural network \cite{Kingma2018}, also known as reversible flows, and a linear time-invariant dynamic system. The \gls{ile} handles nonlinear motion in the pixel space and scales better to longer-term predictions compared to adversarial models~\cite{Mathieu2016}.

While previous variational approaches~\cite{Denton2018,Lee2018} focused on predicting a single frame of low resolution in restricted, predictable or simulated datasets, Hu \textit{et al.}~\cite{Hu2020} jointly predict full-frame ego-motion, static scene, and object dynamics on complex real-world urban driving. Featuring a novel spatio-temporal module, their five-component architecture learns rich representations that incorporate both local and global spatio-temporal context. Authors validated the model on predicting semantic segmentation, depth and optical flow, two seconds in the future outperforming existing spatio-temporal architectures. However, no performance comparison with~\cite{Denton2018,Lee2018} has been carried out.
\section{Performance Evaluation}
\label{sec:performance_eval}
This section presents the results of the previously analyzed video prediction models on the most popular datasets on the basis of the metrics described below. 

\subsection{Metrics and Evaluation Protocols}
\label{subsec:metrics}
For a fair evaluation of video prediction systems, multiple aspects in the prediction have to be addressed such as whether the predicted sequences look realistic, are plausible and cover all possible outcomes. To the best of our knowledge, there are no evaluation protocols and metrics that evaluate the predictions by fulfilling simultaneously all these aspects.

The most widely used evaluation protocols for video prediction rely on image similarity-based metrics such as, \acrfull{mse}, \gls{ssim} \cite{Wang2004}, and \gls{psnr}. However, evaluating a prediction according to the mismatch between its visual appearance and the ground truth is not always reliable. In practice, these metrics penalize all predictions that deviate from the ground truth. In other words, they prefer blurry predictions nearly accommodating the exact ground truth than sharper and plausible but imperfect generations~\cite{Zhang2018a,Lee2018,Castrejon2019}. Pixel-wise metrics do not always reflect how accurate a model captured video scene dynamics and their temporal variability. In addition, the success of a metric is influenced by the loss function used to train the model. For instance, the models trained with \gls{mse} loss function would obviously perform well on \gls{mse} metric, but also on \gls{psnr} metric as it is based on \gls{mse}. Suffering from similar problems, \gls{ssim} measures the similarity between two images, from $-1$ (very dissimilar) to $+1$ (the same image). As a difference, it measures similarities on image patches instead of performing pixel-wise comparison. These metrics are easily fooled by learning to match the background in predictions. To address this issue, Mathieu \textit{et al.}~\cite{Mathieu2016} evaluated the predictions only on the dynamic parts of the sequence, avoiding background influence.

As the pixel space is multimodal and highly-dimensional, it is challenging to evaluate how accurately a prediction sequence covers the full distribution of possible outcomes. Addressing this issue, some probabilistic approaches~\cite{Castrejon2019,Denton2018,Lee2018} adopted a different evaluation protocol to assess prediction coverage. Basically, they sample multiple random predictions and then they search for the best match with the ground truth sequence. Finally, they report the best match using common metrics. This represents the most common evaluation protocol for probabilistic video prediction. Other methods~\cite{Castrejon2019,Jin2020,Shouno2020} also reported results using: \gls{lpips}~\cite{Zhang2018a} as a perceptual metric comparing \gls{cnn} features, or \gls{fvd}~\cite{Unterthiner2018} to measure sample realism by comparing underlying distributions of predictions and ground truth. Moreover, Lee \textit{et al.}~\cite{Lee2018} used the VGG Cosine Similarity metric that performs cosine similarity to the features extracted with the VGGnet~\cite{Simonyan2015} from the predictions.

Some other alternative metrics include the inception score~\cite{Salimans2016} introduced to deal with \glspl{gan} mode collapse problem by measuring the diversity of generated samples; perceptual similarity metrics, such as DeePSiM~\cite{Dosovitskiy2016}; measuring sharpness based on difference of gradients~\cite{Mathieu2016}; Parzen window~\cite{Breuleux2011}, yet deficient for high-dimensional images; and the \gls{log}~\cite{Hildreth1980,Denton2015} used in~\cite{Kwon2019}. In the semantic segmentation space, authors used the popular \gls{iou} metric. Inception score was also widely used to report results on different methods~\cite{Walker2017,Vondrick2016,Denton2017,Villegas2017a}. Differently, on the basis of the \Gls{epva} model~\cite{Wichers2018} a quantitative evaluation was performed, based on the confidence of an external method trained to identify whether the generated video contains a recognizable person. While some authors~\cite{Mathieu2016,Luc2017,Terwilliger2019} evaluated the performance only on the dynamic parts of the image, other directly opted for visual human evaluation through \gls{amt} workers, without a direct quantitative evaluation.

\subsection{Results}
\begin{table}[tbp] 
    \centering
	\ra{1.12}
	\caption{Results on M-MNIST (Moving MNIST). Predicting the next $y$ frames from $x$ context frames ($x\rightarrow y$). $\dag$ results reported by Oliu \textit{et al.}\cite{Oliu2018}, $\ddag$ results reported by Wang \textit{et al.}\cite{Wang2019b}, $\ast$ results reported by Wang \textit{et al.} \cite{Wang2017}, $\triangleleft$ results reported by Wang \textit{et al.} \cite{Wang2018}. \textbf{\gls{mse}} represents per-pixel average \gls{mse} $(10^{-3})$. \textbf{\gls{mse}}$\mathbf{\diamond}$ represents per-frame error. }
	\label{table:results_moving_mnist}
	\resizebox{\linewidth}{!}{
		\begin{tabular} {@{}lccccccc@{}} 
			\toprule
			& \multicolumn{5}{c}{\textbf{M-MNIST}} & \multicolumn{2}{c}{\textbf{M-MNIST}} \\ 
			& \multicolumn{5}{c}{$(10\rightarrow10)$} & \multicolumn{2}{c}{$(10\rightarrow30)$} \\
			\cmidrule{2-6} \cmidrule(l){7-8} 
			\textbf{method} & \textbf{\gls{mse}} & \textbf{\gls{mse}$\mathbf{\diamond}$} & \textbf{\gls{ssim}} & \textbf{\gls{psnr}} & \textbf{\gls{ce}} & \textbf{\gls{mse}$\mathbf{\diamond}$} & \textbf{\gls{ssim}} \\
			\midrule
			BeyondMSE \cite{Mathieu2016} & $27.48\dag$ & $122.6\ast$ & $0.713\ast$ & $15.969\dag$ & - & - & - \\
			Srivastava \textit{et al.} \cite{Srivastava2015} & $17.37\dag$ & $118.3\ast$ & $0.690\ast$ & $18.183\dag$ & $341.2$ & $180.1\triangleleft$ & $0.583\triangleleft$\\ 
			Shi \textit{et al.} \cite{Shi2015} & - & $96.5\ddag$ & $0.713\ddag$ & - & $367.2\ast$ & $156.2\triangleleft$ & $0.597\triangleleft$\\
			\gls{dfn} \cite{Brabandere2016} & - & $89.0\ddag$ & $0.726\ddag$ & - & $285.2$ & $149.5\triangleleft$ & $0.601\triangleleft$\\
			\gls{cdna} \cite{Finn2016} & - & $84.2\ddag$ & $0.728\ddag$ & - & $346.6\ast$ & $142.3\triangleleft$ & $0.609\triangleleft$\\
			\acrshort{vln} \cite{Cricri2016} & - & - & - & - & $187.7$ \\
			Patraucean \textit{et al.} \cite{Patraucean2015} & $43.9$ & - & - & - & $179.8$ & - & -\\
			\gls{mcnet} \cite{Villegas2017a}$\dag$ & $42.54$ & - & - & $13.857$ & - & - & -\\
			\acrshort{rln} \cite{Premont-Schwarz2017}$\dag$ & $42.54$ & - & - & $13.857$ & - & - & -\\
			\gls{prednet} \cite{Lotter2017}$\dag$ & $41.61$ & - & - & $13.968$ & - & - & - \\
			\acrshort{frnn} \cite{Oliu2018} & $\mathbf{9.47}$ & $68.4\ddag$ & $0.819\ddag$ & $\mathbf{21.386}$ & - & - & -\\
			PredRNN \cite{Wang2017} & - & $56.8$ & $0.867$ & - & $97.0$ & - & -\\
			\acrshort{vpn} \cite{Kalchbrenner2016} & - & $64.1\ddag$ & $0.870\ddag$ & - & $\mathbf{87.6}$ & $129.6\triangleleft$ & $0.620\triangleleft$\\
			Znet \cite{Zhang2019} & - & $50.5$ & $0.877$ & - & - & - & -\\
			PredRNN++ \cite{Wang2018} & - & $46.5$ & $0.898$ & - & - & $\mathbf{91.1}$ & $\mathbf{0.733}$\\
			\gls{e3dlstm} \cite{Wang2019b} & - & $41.3$ & $0.910$ & - & - & - & -\\
			\gls{crevnet} \cite{Yu2020} & - & $\mathbf{22.3}$ & $\mathbf{0.949}$ & - & - & - & -\\
			\bottomrule
	\end{tabular}}
\end{table}
In this section we report the quantitative results of the most relevant methods reviewed in the previous sections. To achieve a wide comparison, we limited the quantitative results to the most common metrics and datasets. We have distributed the results in different tables, given the large variation in the evaluation protocols of the video prediction models. 

Many authors evaluated their methods on the Moving MNIST synthetic environment. Although it represents a restricted and quasi-deterministic scenario, long-term predictions are still challenging. The black and homogeneous background induce methods to accurately extrapolate black frames and vanish the predicted digits in the long-term horizon. Under this configuration, the \gls{crevnet} model demonstrated a leap over the previous state of the art. As the second best, the \gls{e3dlstm} network reported stable errors in both short-term and longer-term predictions showing the advantages of their memory attention mechanism. It also reported the second best results on the KTH dataset, after \cite{Jin2020} which achieved the best overall performance and demonstrated quality predictions on natural videos.

\begin{table}[tbp] 
    \centering
	\ra{1.11}
	\caption{Results on KTH dataset. Predicting the next $y$ frames from $x$ context frames ($x\rightarrow y$). $\dag$ results reported by Oliu \textit{et al.}\cite{Oliu2018}, $\ddag$ results reported by Wang \textit{et al.} \cite{Wang2019b}, $\ast$ results reported by Zhang \textit{et al.} \cite{Zhang2019}, $\triangleleft$ results reported by Jin \textit{et al.} \cite{Jin2020}. Per-pixel average \gls{mse} $(10^{-3})$. Best results are represented in bold.}
	\label{table:results_kth}
	\resizebox{\linewidth}{!}{
		\begin{tabular} {@{}lcccccc@{}} 
			\toprule
			& \multicolumn{2}{c}{\textbf{KTH}} &\multicolumn{2}{c}{\textbf{KTH}} & \multicolumn{2}{c}{\textbf{KTH}} \\ 
			& \multicolumn{2}{c}{\textbf{($10\rightarrow10$)}} &\multicolumn{2}{c}{\textbf{($10\rightarrow20$)}} & \multicolumn{2}{c}{\textbf{($10\rightarrow40$)}} \\ 
			\cmidrule{2-3} \cmidrule(lr){4-5} \cmidrule{6-7}
			\textbf{method} & \textbf{\gls{mse}} & \textbf{\gls{psnr}} & \textbf{\gls{ssim}} & \textbf{\gls{psnr}} & \textbf{\gls{ssim}} & \textbf{\gls{psnr}} \\
			\midrule
			Srivastava \textit{et al.} \cite{Srivastava2015}$\dag$ & $9.95$ & $21.22$ & - & - & - & - \\
			\gls{prednet} \cite{Lotter2017}$\dag$ & $3.09$ & $28.42$ & - & - & - & - \\
			BeyondMSE \cite{Mathieu2016}$\dag$ & $1.80$ & $29.34$ & - & - & - & - \\
			\acrshort{frnn} \cite{Oliu2018} & $1.75$ & $29.299$ & $0.771\triangleleft$ & $26.12\triangleleft$ & $0.678\triangleleft$ & $23.77\triangleleft$ \\
			\gls{mcnet} \cite{Villegas2017a} & $1.65\dag$ & $30.95\dag$ & $0.804\ddag$ & $25.95\ddag$ & $0.73\triangleleft$& $23.89\triangleleft$ \\
			\acrshort{rln} \cite{Premont-Schwarz2017}$\dag$ & $\mathbf{1.39}$ & $\mathbf{31.27}$ & - & - & - & - \\
			Shi \textit{et al.} \cite{Shi2015}$\ddag$ & - & - & $0.712$ & $23.58$ & $0.639$ & $22.85$ \\
			\gls{savp}\cite{Lee2018}$\triangleleft$ & - & - & $0.746$ & $25.38$ & $0.701$ & $23.97$ \\
			\acrshort{vpn} \cite{Kalchbrenner2016}$\ast$ & - & - & $0.746$ & $23.76$ & - & - \\
			\gls{dfn} \cite{Brabandere2016}$\ddag$ & - & - & $0.794$ & $27.26$ & $0.652$ & $23.01$ \\
			\acrshort{frnn} \cite{Oliu2018}$\ddag$ & - & - & $0.771$ & $26.12$ & $0.678$ & $23.77$ \\
			Znet \cite{Zhang2019} & - & - & $0.817$ & $27.58$ & - & - \\
			\gls{sv2p} invariant\cite{Babaeizadeh2018}$\triangleleft$ & - & - & $0.826$ & $27.56$ & $0.778$ & $25.92$ \\
			\gls{sv2p} variant\cite{Babaeizadeh2018}$\triangleleft$ & - & - & $0.838$ & $27.79$ & $0.789$ & $26.12$ \\
			PredRNN \cite{Wang2017} & - & - & $0.839$ & $27.55$ & $0.703\ddag$ & $24.16\ddag$ \\
			VarNet \cite{Jin2018}$\triangleleft$ & - & - & $0.843$ & $28.48$ & $0.739$ & $25.37$ \\
			\gls{savp}-VAE\cite{Lee2018}$\triangleleft$ & - & - & $0.852$ & $27.77$ & $0.811$ & $26.18$ \\
			PredRNN++ \cite{Wang2018} & - & - & $0.865$ & $28.47$ & $0.741\ddag$ & $25.21\ddag$ \\
			MSNET \cite{Lee2018a} & - & - & $0.876$ & $27.08$ & - & - \\
			\gls{e3dlstm} \cite{Wang2019b} & - & - & $0.879$ & $29.31$ & $0.810$ & $27.24$ \\
			Jin \textit{et al.}\cite{Jin2020} & - & - & $\mathbf{0.893}$ & $\mathbf{29.85}$ & $\mathbf{0.851}$ & $\mathbf{27.56}$ \\
			\bottomrule
	\end{tabular}}
\end{table}

Performing short-term predictions in the KTH dataset, the \gls{rln} outperformed \gls{mcnet} and \gls{frnn} by a slight margin. The \gls{rln} architecture draws similarities with \gls{frnn}, except that the former uses bridge connections and the latter, state sharing that improves memory consumption. On the Moving MNIST and UCF101 datasets, \gls{frnn} outperformed \gls{rln}. Other interesting methods to highlight are PredRNN and PredRNN++, both providing close results to \gls{e3dlstm}. State-of-the-art results using different metrics were reported on Caltech Pedestrian by Kwon \textit{et al.}~\cite{Kwon2019}, \gls{crevnet}~\cite{Yu2020}, and Jin \textit{et al.}~\cite{Jin2020}. The former, by taking advantage of its retrospective prediction scheme, was also the overall winner on the UCF-101 dataset meanwhile the latter outperformed previous methods on the BAIR Push dataset.

On the one hand, some approaches have been evaluated on other datasets: SDC-Net~\cite{Reda2018} outperformed~\cite{Mathieu2016,Villegas2017a} on YouTube8M, \gls{trivdganfp} outperformed~\cite{Clark2019,Weissenborn2020} on Kinetics-600 test set \cite{Carreira2018}, \gls{e3dlstm} compared their method with~\cite{Kalchbrenner2016,Oliu2018,Wang2017,Wang2018} on the TaxiBJ dataset~\cite{Zhang2017a}, and \gls{crevnet}~\cite{Yu2020} on Traffic4cast~\cite{traffic4cast}. On the other hand, some explored out-of-domain tasks~\cite{Shi2015,Wang2019b,Brabandere2016,Vondrick2017,Yu2020} (see ood column in Table \ref{table:methods}). 

\begin{table} \centering
	\ra{1.11}
	\caption{Results on Caltech Pedestrian. Predicting the next $y$ frames from $x$ context frames ($x\rightarrow y$). $\dag$ reported by Kwon \textit{et al.} \cite{Kwon2019}, $\ddag$ reported by Reda \textit{et al.} \cite{Reda2018}, $\ast$ reported by Gao \textit{et al.} \cite{Gao2019}, $\triangleleft$ reported by Jin \textit{et al.} \cite{Jin2020}. Per-pixel average \gls{mse} $(10^{-3})$. Best results are represented in bold.}
	\label{table:results_caltech}
	\resizebox{0.75\linewidth}{!}{
	\begin{tabular} {@{}lcccc@{}} 
		\toprule
		& \multicolumn{4}{c}{\textbf{Caltech Pedestrian}} \\ 
		& \multicolumn{4}{c}{\textbf{($10\rightarrow1$)}} \\ 
		\cmidrule{2-5} 
		\textbf{method} & \textbf{\gls{mse}} & \textbf{\gls{ssim}} & \textbf{\gls{psnr}} & \textbf{\gls{lpips}}\\
		\midrule
		BeyondMSE \cite{Mathieu2016}$\ddag$ & $3.42$ & $0.847$ & - & - \\ 
		\gls{mcnet} \cite{Villegas2017a}$\ddag$ & $2.50$ & $0.879$ & - & - \\
		\gls{dvf} \cite{Liu2017}$\ast$ & - & $0.897$ & $26.2$ & $5.57\triangleleft$\\
		Dual-GAN \cite{Liang2017} & $2.41$ & $0.899$ & - & -\\
		CtrlGen \cite{Hao2018}$\ast$ & - & $0.900$ & $26.5$ & $6.38\triangleleft$ \\ 
		\gls{prednet} \cite{Lotter2017}$\dag$ & $2.42$ & $0.905$ & $27.6$ & $7.47\triangleleft$\\ 
		ContextVP \cite{Byeon2018} & $1.94$ & $0.921$ & $28.7$ & $6.03\triangleleft$\\ 
		GAN-VGG \cite{Shouno2020} & - & $0.916$ & - & $3.61$ \\
		G-VGG \cite{Shouno2020} & - & $0.917$ & - & $\mathbf{3.52}$ \\
		SDC-Net \cite{Reda2018} & $1.62$ & $0.918$ & - & -\\ 
		Kwon et al. \cite{Kwon2019} & $\mathbf{1.61}$ & $0.919$ & $29.2$ & -\\ 
		DPG \cite{Gao2019} & $-$ & $0.923$ & $28.2$ & $5.04\triangleleft$\\ 
		G-MAE \cite{Shouno2020} & - & $0.923$ & - & $4.30$ \\
		GAN-MAE \cite{Shouno2020} & - & $0.923$ & - & $4.09$ \\
		\gls{crevnet} \cite{Yu2020} & - & $0.925$ & $\mathbf{29.3}$ & - \\
		Jin \textit{et al.}\cite{Jin2020} & - & $\mathbf{0.927}$ & $29.1$ & $5.89$ \\
		\bottomrule
	\end{tabular}}
\end{table}
\begin{table}[tbp] 
    \centering
	\ra{1.11}
	\caption{Results on UCF-101 dataset. Predicting the next $x$ frames from $y$ context frames ($x\rightarrow y$). $\dag$ results reported by Oliu \textit{et al.}\cite{Oliu2018}. Per-pixel average \gls{mse} $(10^{-3})$. Best results are represented in bold.}
	\label{table:results_ucf101}
	\resizebox{0.9\linewidth}{!}{
		\begin{tabular} {@{}lcccccc@{}} 
			\toprule
			& \multicolumn{2}{c}{\textbf{UCF-101}} & \multicolumn{3}{c}{\textbf{UCF-101}}\\ 
			& \multicolumn{2}{c}{\textbf{($10\rightarrow10$)}} & \multicolumn{3}{c}{\textbf{($4\rightarrow1$)}}\\ 
			\cmidrule{2-3} \cmidrule(l){4-6} 
			\textbf{method} & \textbf{\gls{mse}} & \textbf{\gls{psnr}} & \textbf{\gls{mse}} & \textbf{\gls{ssim}} & \textbf{\gls{psnr}} \\
			\midrule
			Srivastava \textit{et al.} \cite{Srivastava2015}$\dag$ & $148.66$ & $10.02$ & - & - & - \\
			\gls{prednet} \cite{Lotter2017}$\dag$ & $15.50$ & $19.87$ & - & - & -\\
			BeyondMSE \cite{Mathieu2016}$\dag$ & $9.26$ & $22.78$ & - & - & -\\ 
			\gls{mcnet} \cite{Villegas2017a}& $9.40\dag$ & $23.46\dag$ & - & $0.91$ & $31.0$ \\
			\acrshort{rln} \cite{Premont-Schwarz2017}$\dag$ & $9.18$ & $23.56$ & - & - & -\\
			\acrshort{frnn} \cite{Oliu2018} & $\mathbf{9.08}$ & $\mathbf{23.87}$ & - & - & -\\
			BeyondMSE \cite{Mathieu2016} & - & - & - & $0.92$ & $32$ \\ 
			Dual-GAN \cite{Liang2017} & - & - & - & $\mathbf{0.94}$ & $30.5$ \\
			\gls{dvf} \cite{Liu2017} & - & - & - & $\mathbf{0.94}$ & $33.4$ \\ 
			ContextVP \cite{Byeon2018} & - & - & - & $0.92$ & $34.9$ \\
			Kwon et al. \cite{Kwon2019} & - & - & $\mathbf{1.37}$ & $\mathbf{0.94}$ & $\mathbf{35.0}$ \\
			\bottomrule
	\end{tabular}}
\end{table}

\subsubsection{Results on Probabilistic Approaches}
Video prediction probabilistic methods have been mainly evaluated on the Stochastic Moving MNIST, Bair Push and Cityscapes datasets. Different from the original Moving MNIST dataset, the stochastic version includes uncertain digit trajectories, i.e. the digits bounce off the border with a random new direction. On this dataset, both versions of Castrejon \textit{et al.} models (1L, without a hierarchy of latents, and 3L with a 3-level hierarchy of latents) outperform \gls{svg} by a large margin. On the Bair Push dataset, \gls{savp} reported sharper and more realistic-looking predictions than \gls{svg} which suffer of blurriness. However, both models were outperformed by~\cite{Castrejon2019} as well on the Cityscapes dataset. The model based on a 3-level hierarchy of latents~\cite{Castrejon2019} outperform previous works on all three datasets, showing the advantages of the extra expressiveness of this model.

\begin{table} 
    \centering
	\ra{1.12}
	\caption{Results on SM-MNIST (Stochastic Moving MNIST), BAIR Push and Cityscapes datasets. $\dag$ results reported by Castrejon \textit{et al.} \cite{Castrejon2019}. $\ddag$ results reported by Jin \textit{et al.}\cite{Jin2020}.}
	\label{table:results_stochastic}
	\resizebox{\linewidth}{!}{
		\begin{tabular} {@{}lccccccc@{}} 
			\toprule
			& \multicolumn{2}{c}{\textbf{SM-MNIST}} & \multicolumn{3}{c}{\textbf{BAIR Push}} & \multicolumn{2}{c}{\textbf{Cityscapes}} \\ 
			& \multicolumn{2}{c}{\textbf{($5\rightarrow10$)}} & \multicolumn{3}{c}{\textbf{($2\rightarrow28$)}} & \multicolumn{2}{c}{\textbf{($2\rightarrow28$)}} \\ 
			\cmidrule(lr){2-3} \cmidrule(lr){4-6} \cmidrule(lr){7-8} 
			\textbf{method} & \textbf{\acrshort{fvd}} & \textbf{\gls{ssim}} & \textbf{\acrshort{fvd}} & \textbf{\gls{ssim}} & \textbf{\acrshort{psnr}} & \textbf{\acrshort{fvd}} & \textbf{\gls{ssim}} \\
			\midrule
			\gls{svg}\cite{Denton2018} & $90.81\dag$ & $0.688\dag$ & $256.62\dag$ & $0.816\dag$ & $17.72\ddag$ & $1300.26\dag$ & $0.574\dag$ \\
			\gls{savp}\cite{Lee2018} & - & - & $143.43\dag$ & $0.795\dag$ & $18.42\ddag$ & - & - \\
			\gls{savp}-VAE\cite{Lee2018} & - & - & - & $0.815\ddag$ & $19.09\ddag$ & - & - \\
			\gls{sv2p} inv.\cite{Babaeizadeh2018}$\ddag$ & - & - & - & $0.817$ & $20.36$ & - & - \\
			vRNN 1L \cite{Castrejon2019} & $63.81$ & $\mathbf{0.763}$ & $149.22$ & $0.829$ & - & $682.08$ & $0.609$ \\
			vRNN 3L \cite{Castrejon2019} & $\mathbf{57.17}$ & $0.760$ & $\mathbf{143.40}$ & $0.822$ & - & $\mathbf{567.51}$ & $\mathbf{0.628}$ \\
			Jin \textit{et al.}\cite{Jin2020} & - & - & - & $\mathbf{0.844}$ & $\mathbf{21.02}$ & - & - \\
			\bottomrule
	\end{tabular}}
\end{table}

\subsubsection{Results on the High-level Prediction Space}
\begin{table} \centering
	\ra{1.11}
	\caption{Results on Cityscapes dataset. Predicting the next $y$ time-steps of semantic segmented frames from 4 context frames ($4\rightarrow y$). $\ddag$ \gls{iou} results on eight moving objects classes. $\dag$ results reported by Chiu \textit{et al.} \cite{Chiu2019}}
	\label{table:results_segmenation}
	\resizebox{0.9\linewidth}{!}{
		\begin{tabular} {@{}lcccc@{}} 
			\toprule
			& \multicolumn{4}{c}{\textbf{Cityscapes}} \\ 
			& $(4\rightarrow1)$ & $(4\rightarrow3)$ & $(4\rightarrow9)$ & $(4\rightarrow10)$ \\
			\cmidrule{2-5} 
			\textbf{method} & \textbf{\gls{iou}} & \textbf{\gls{iou}} & \textbf{\gls{iou}} & \textbf{\gls{iou}}\\
			\midrule
			S2S \cite{Luc2017}$\ddag$ & - & $55.3$ & $40.8$ & -\\ 
			S2S-maskRCNN \cite{Luc2018}$\ddag$ & - & $55.4$ & $42.4$ & -\\
			S2S \cite{Luc2017} & $62.60\ddag$ & $59.4$ & $47.8$ & -\\ 
			Nabavi \textit{et al.} \cite{Nabavi2018} &$71.37$ & $60.06$ & - & - \\ 
			F2F \cite{Luc2018} & - & $61.2$ & $41.2$ & -\\
			Vora \textit{et al.} \cite{Vora2018} & - & $61.47$ & $45.4$ & - \\
			S2S-Res101-FCN \cite{Jin2017a} & - & $62.6$ & - & $50.8$\\
			Terwilliger \textit{et al.} \cite{Terwilliger2019}$\ddag$ & - & $65.1$ & $46.3$ & - \\ 
			Chiu \textit{et al.} \cite{Chiu2019} & $72.43$ & $65.53$ & $50.52$ \\ 
			Jin \textit{et al.} \cite{Jin2017a} & - & $66.1$ & - & $\mathbf{53.9}$\\ 
			Bayes-WD-SL \cite{Bhattacharyya2019} & $\mathbf{75.3}$ & $66.7$ & $\mathbf{52.5}$ & - \\ 
			Terwilliger \textit{et al.} \cite{Terwilliger2019} & $73.2$ & $\mathbf{67.1}$ & $51.5$ & $52.5$ \\ 
			\bottomrule
	\end{tabular}}
\end{table}

Most of the methods have chosen the semantic segmentation space to make predictions. Although they relied on different datasets for training, performance results were mostly reported on the Cityscapes dataset using the \gls{iou} metric. Authors explored short-term (next-frame prediction), mid-term (+3 time steps in the future) and long-term (up to +10 time step in the future) predictions. On the semantic segmentation prediction space, Bayes-WD-SL~\cite{Bhattacharyya2019}, the model proposed by Terwilliger \textit{et al.}~\cite{Terwilliger2019}, and Jin \textit{et al.}~\cite{Jin2017} reported the best results. Among these methods, it is noteworthy that Bayes-WD-SL was the only one to explore prediction diversity on the basis of a Bayesian formulation.

In the instance segmentation space, the F2F pioneering method~\cite{Luc2018} was outperformed by Sun \textit{et al.}~\cite{Sun2019} on short and mid-term predictions using the AP50 and AP evaluation metrics. On the other hand, in the keypoint coordinate space, the seminal model of Minderer \textit{et al.}~\cite{Minderer2019} qualitatively outperforms \gls{svg} \cite{Denton2018}, \gls{savp} \cite{Lee2018} and \gls{epva} \cite{Wichers2018}, yet pixel-wise metrics reported similar results. In the human pose space, Tang \textit{et al.}~\cite{Tang2019} by regressing future frames from human pose predictions outperformed \gls{savp}~\cite{Lee2018}, \gls{mcnet}~\cite{Villegas2017a} and \cite{Villegas2017} on the basis of the \gls{psnr} and \gls{ssim} metrics on the Penn Action and J-HMDB \cite{Jhuang2013} datasets. 
\section{Discussion}
\label{sec:discussion}
The video prediction literature ranges from a direct synthesis of future pixel intensities, to complex probabilistic models addressing prediction uncertainty. The range between these approaches consists of methods that try to factorize or narrow the prediction space. Simplifying the prediction task has been a natural evolution of video prediction models, influenced by several open research challenges discussed below. Due to the curse of dimensionality and the inherent pixel variability, developing a robust prediction based on raw pixel intensities is overly-complicated. This often leads to the regression-to-the-mean problem, visually represented as blurriness. Making parametric models larger would improve the quality of predictions, yet this is currently incompatible with high-resolution predictions due to memory constraints. Transformation-based approaches propagate pixels from previous frames based on estimated flow maps. In this case, prediction quality is directly influenced by the accuracy of the estimated flow. Similarly, the prediction in a high-level space is mostly conditioned by the quality of some extra supervisory signals such as semantic maps and human poses, to name a few. Erroneous supervision signals would harm prediction quality.

Analyzing the impact of the inductive bias on the performance of a video prediction model, Villegas \textit{et al.}~\cite{Villegas2019} demonstrated the maximization of the \gls{svg} model~\cite{Denton2018} performance with minimal inductive bias (e.g. segmentation or instance maps, optical flow, adversarial losses, etc.) by increasing progressively the scale of computation. A common assumption when addressing the prediction task in a high-level feature space, is the direct improvement of long-term predictions as a result of simplifying the prediction space. Even if the complexity of the prediction space is reduced, it is still multimodal when dealing with natural videos. For instance, when it comes to long-term predictions in the semantic segmentation space, most of the models reported predictions only up to ten time steps into the future. This directly suggests that the choice of the prediction space is still an unsolved problem. Finding a trade-off between the complexity of the prediction space and the output quality is challenging. An overly-simplified representation could limit the prediction on complex data such as natural videos. Although abstract predictions suffice for many of the decision-making systems based on visual reasoning, prediction in pixel space is still being addressed. 

From the analysis performed in this review and in line with the conclusions extracted from~\cite{Villegas2019} we state that: (1)~including recurrent connections and stochasticity in a video prediction model generally lead to improved performance; (2)~increasing model capacity while maintaining a low inductive bias also improves prediction performance; (3)~multi-step predictions conditioned by previously generated outputs are prone to accumulate errors, diverging from the ground truth when addressing long-term horizons; (4)~authors predicted further in the future without relying on high-level feature spaces; (5)~combining pixel-wise losses with adversarial training somewhat mitigates the regression-to-the-mean issue.

\subsection{Research Challenges}
Despite the wealth of currently existing video prediction approaches and the significant progress made in this field, there is still room to improve state-of-the-art algorithms. To foster progress, open research challenges must be clearly identified and disentangled. So far in this review, we have already discussed about: (1)~the importance of spatio-temporal correlations as a self-supervisory signal for predictive models; (2)~how to deal with future uncertainty and model the underlying multimodal distributions of natural videos; (3)~the over-complicated task of learning meaningful representations and deal with the curse of dimensionality; (4)~pixel-wise loss functions and blurry results when dealing with equally probable outcomes, i.e. probabilistic environments. These issues define the open research challenges in video prediction.

Currently existing methods are limited to short-term horizons. While frames in the immediate future are extrapolated with high accuracy, in the long term horizon the prediction problem becomes multimodal by nature. Initial solutions consisted on conditioning the prediction on previously predicted frames. However, these autoregressive models tend to accumulate prediction errors that progressively diverge the generated prediction from the expected outcome. On the other hand, due to memory issues, there is a lack of resolution in predictions. Authors tried to address this issue by composing the full-resolution image from small predicted patches. However, as the results are not convincing because of the annoying tilling effect, most of the available models are still limited to low-resolution predictions. In addition to the lack of resolution and long-term predictions, models are still prone to the regress-to-the-mean problem that consists on averaging the output frame to accommodate multiple equally probable outcomes. This is directly related to the pixel-wise loss functions, that focus the learning process on the visual appearance. The choice of the loss function is an open research problem with a direct influence on the prediction quality. Finally, the lack of reliable and fair evaluation models makes the qualitative evaluation of video prediction challenging and represents another potential open problem.

\subsection{Future Directions}
\label{subsec:future}
Based on the reviewed research identifying the state-of-the-art video prediction methods, we present some future promising research directions.

\vspace*{0.15cm}\noindent\textbf{Consider alternative loss functions}: Pixel-wise loss functions are widely used in the video prediction task, causing blurry predictions when dealing with uncontrolled environments or long-term horizon. In this regard, great efforts have been made in the literature for identifying more suitable loss functions for the prediction task. However, despite the existing wide spectrum of loss functions, most models still blindly rely on deterministic loss functions.

\vspace*{0.15cm}\noindent\textbf{Alternatives to \glspl{rnn}}: Currently, \glspl{rnn} are still widely used in this field to model temporal dependencies, and achieved state-of-the-art results on different benchmarks \cite{Oliu2018,Wang2018,Wang2017,Wang2019b}. Nevertheless, some methods also relied on 3D convolutions to further enhance video prediction~\cite{Wang2019b,Chiu2019} representing a promising avenue.

\vspace*{0.15cm}\noindent\textbf{Use synthetically generated videos}: Simplifying the prediction is a current trend in the video prediction literature. A vast amount of video prediction models explored higher-level features spaces to reformulate the prediction task into a more tractable problem. However, this mostly conditions the prediction to the accuracy of an external source of supervision such as optical flow, human pose, pre-activations (percepts) extracted from supervised networks, and more. However, this issue could be alleviated by taking advantage of existing fully-annotated and photorealistic synthetic datasets or by using data generation tools. Video prediction in photorealistic synthetic scenarios has not been explored in the literature.

\vspace*{0.15cm}\noindent\textbf{Evaluation metrics}: Since the most widely used evaluation protocols for video prediction rely on image similarity-based metrics, the need for fairer evaluation metrics is imminent. A fair metric should not penalize predictions that deviate from the ground truth at the pixel level, if their content represents a plausible future prediction in a higher level, i.e., the dynamics of the scene correspond to the reality of the labels. In this regard, some methods evaluate the similarity between distributions or at a higher-level. However, there is still room for improvement in the evaluation protocols for video prediction and generation~\cite{Theis2016}.
\section{Conclusion}
\label{sec:conclusion}
In this review, after reformulating the predictive learning paradigm in the context of video prediction, we have closely reviewed the fundamentals on which it is based: exploiting the time dimension of videos, dealing with stochasticity, and the importance of the loss functions in the learning process. Moreover, an analysis of the backbone deep learning-based architectures for this task was performed in order to provide the reader the necessary background knowledge. The core of this study encompasses the analysis and classification of more than 50 methods and the datasets they have used. Methods were analyzed from three perspectives: method description, contribution over the previous works and performance results. They have also been classified according to a proposed taxonomy based on their main contribution. In addition, we have presented a comparative summary of the datasets and methods in tabular form so as the reader, at a glance, could identify low-level details. In the end, we have discussed the performance results on the most popular datasets and metrics to finally provide useful insight in shape of future research directions and open problems. In conclusion, video prediction is a promising avenue for the self-supervised learning of rich spatio-temporal correlations to provide prediction capabilities to existing intelligent decision-making systems. While great strides have been made, there is still room for improvement in video prediction using deep learning techniques.

\ifCLASSOPTIONcompsoc
  \section*{Acknowledgments}
\else
  \section*{Acknowledgment}
\fi

This work has been funded by the Spanish Government PID2019-104818RB-I00 grant for the MoDeaAS project. This work has also been supported by two Spanish national grants for PhD studies, FPU17/00166, and ACIF/2018/197 respectively. Experiments were made possible by a generous hardware donation from NVIDIA.

\ifCLASSOPTIONcaptionsoff
  \newpage
\fi



%
\bibliographystyle{IEEEtran}
\bibliography{references}

%

\begin{IEEEbiography}[{\includegraphics[width=1in,height=1.25in,clip,keepaspectratio]{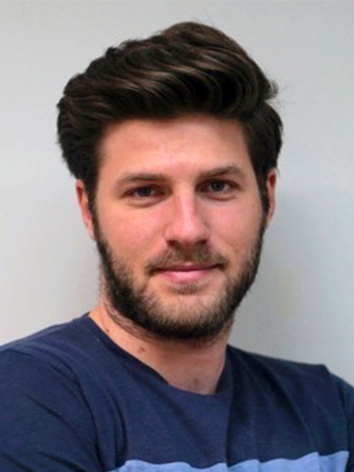}}]{Sergiu Oprea} is a PhD student at the Department of Computer Technology (DTIC), University of Alicante. He received his MSc (Automation and Robotics) and BSc (Computer Science) from the same institution in 2017 and 2015 respectively. His main research interests include video prediction with deep learning, virtual reality, 3D computer vision, and parallel computing on GPUs.
\end{IEEEbiography}
\begin{IEEEbiography}[{\includegraphics[width=1in,height=1.25in,clip,keepaspectratio]{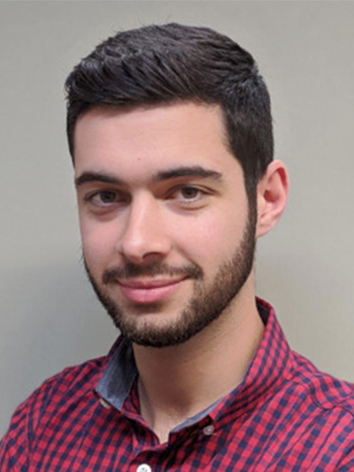}}]{Pablo Martinez Gonzalez} is a PhD student at the Department of Computer Technology (DTIC), University of Alicante. He received his MSc (Computer Graphics, Games and Virtual Reality) and BSc (Computer Science) at the Rey Juan Carlos University and University of Alicante, in 2017 and 2015, respectively. His main research interests include deep learning, virtual reality and parallel computing on GPUs.
\end{IEEEbiography}
\begin{IEEEbiography}[{\includegraphics[width=1in,height=1.25in,clip,keepaspectratio]{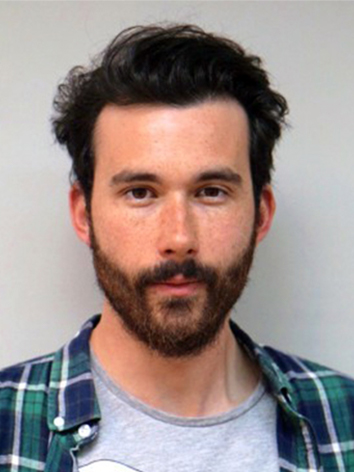}}]{Alberto Garcia Garcia} is a Postdoctoral Researcher at the Institute of Space Sciences (ICE-CSIC, Barcelona) where he leads the efforts in code optimization, machine learning, and parallel computing on the MAGNESIA ERC Consolidator project. He received his PhD (Machine Learning and Computer Vision), MSc (Automation and Robotics) and BSc (Computer Science) from the same institution in 2019, 2016 and 2015 respectively. Previously he was an intern at NVIDIA Research/Engineering, Facebook Reality Labs, and Oculus Core Tech. His main research interests include deep learning (specially convolutional neural networks), virtual reality, 3D computer vision, and parallel computing on GPUs.
\end{IEEEbiography}
\begin{IEEEbiography}[{\includegraphics[width=1in,height=1.25in,clip,keepaspectratio]{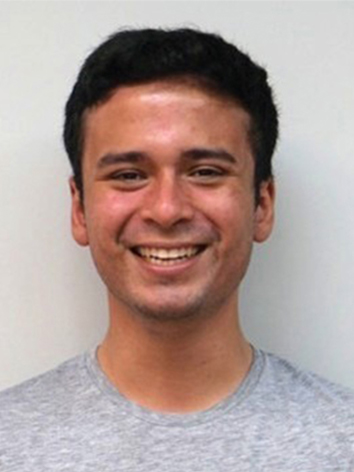}}]{John Alejandro Castro Vargas} is a PhD student at the Department of Computer Technology (DTIC), University of Alicante. He received his MSc (Automation and Robotics) and BSc (Computer Science) from the same institution in 2017 and 2016 respectively. His main research interests include human behavior recognition with deep learning, virtual reality and parallel computing on GPUs.
\end{IEEEbiography}
\begin{IEEEbiography}[{\includegraphics[width=1.0in,height=1.25in,clip,keepaspectratio]{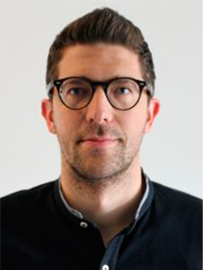}}]{Sergio Orts-Escolano} received a BSc, MSc and PhD in Computer Science from the University of Alicante in 2008, 2010 and 2014 respectively. His research interests include computer vision, assistive robotics, 3D sensors, GPU computing, virtual/augmented reality and deep learning. He has authored +50 publications in top journals and conferences like CVPR, SIGGRAPH, 3DV, BMVC, CVIU, IROS, UIST, RAS, etcetera. He is also a member of European Networks like HiPEAC and Eucog. He has experience as a professor in academia and industry, working as a research scientist for companies such as Google and Microsoft Research.

\end{IEEEbiography}

\begin{IEEEbiography}[{\includegraphics[width=1in,height=1.25in,clip,keepaspectratio]{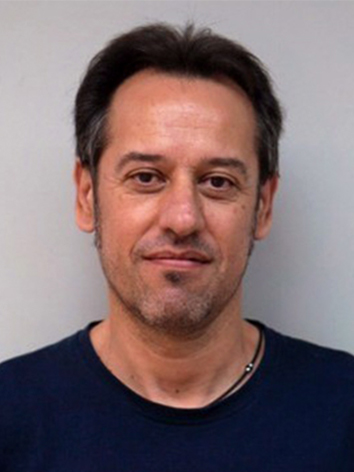}}]{Jose Garcia-Rodriguez}
	received his Ph.D. degree, with specialization in Computer Vision and Neural Networks, from the University of Alicante (Spain). He is currently Full Professor at the Department of Computer Technology of the University of Alicante. His research areas of interest include: computer vision, computational intelligence, machine learning, pattern recognition, robotics, man-machine interfaces, ambient intelligence, computational chemistry, and parallel and multicore architectures.
\end{IEEEbiography}

\begin{IEEEbiography}[{\includegraphics[width=1in,height=1.25in,clip,keepaspectratio]{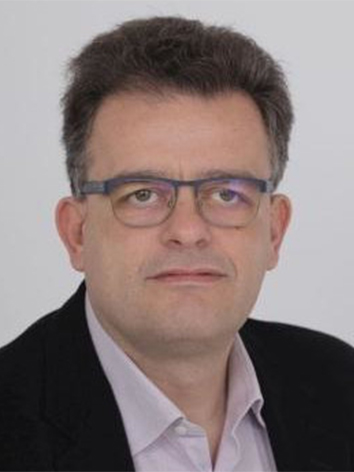}}]{Antonis Argyros} is a professor of computer science at the Computer Science Department, University of Crete and a researcher at the Institute of Computer Science, FORTH, in Heraklion, Crete, Greece. His research interests fall in the areas of computer vision and pattern recognition, with
emphasis on the analysis of humans in images and videos, human pose analysis, recognition of human activities and gestures, 3D computer vision, as well as image motion and tracking. He is
also interested in applications of computer vision in the fields of robotics and smart environments. 
\end{IEEEbiography}

\vfill



\end{document}